\documentclass[10pt,twocolumn,letterpaper]{article}
\usepackage[pagenumbers]{cvpr}
\usepackage[T1]{fontenc}
\usepackage{amsmath,amssymb,amsthm,booktabs,graphicx,placeins,array,multirow,caption,xcolor}
\usepackage{colortbl}
\usepackage{enumitem}
\definecolor{cvprblue}{rgb}{0.21,0.49,0.74}
\usepackage[pagebackref,breaklinks,colorlinks,allcolors=cvprblue]{hyperref}
\usepackage[capitalize]{cleveref}
\makeatletter\expandafter\def\csname ver@eso-pic.sty\endcsname{2023/05/03}\makeatother
\usepackage{pdfpages}
\newcommand{\E}{\mathbb E}
\newcommand{\norm}[1]{\lVert #1\rVert}
\newcommand{\ip}[2]{\langle #1, #2\rangle}
\theoremstyle{plain}
\newtheorem{proposition}{Proposition}
\def\confName{CVPR}
\def\confYear{2027}
\newcommand{\dep}{\mathrm{dep}}
\newcommand{\tb}{\bar t}
\newcommand{\clip}{\mathrm{clip}}
\newcommand{\ciCrop}{256}
\newcommand{\ciRank}{4}
\newcommand{\ciSteps}{1000}
\newcommand{\gsixNUnits}{431}
\newcommand{\gsixBootN}{4000}
\newcommand{\gsixBootLevel}{95}
\newcommand{\gsixNSettings}{5}
\newcommand{\gsixNCameras}{20}
\newcommand{\gsixNPublished}{9}
\newcommand{\gsixrthNPub}{9}
\newcommand{\gsixrthChPuHoldHarmU}{20.3}
\newcommand{\gsixVKTwoIoH}{7.46}
\newcommand{\gsixVKTwoPuH}{16.89}
\newcommand{\gsixVKTwoSdH}{3.55}
\newcommand{\gsixVKTwoNrH}{6.92}
\newcommand{\gsixVKTwoNgH}{3.09}
\newcommand{\gsixVKTwoAllG}{1.923}
\newcommand{\gsixVKTwoAllH}{7.58}
\newcommand{\gsixVFullIoH}{16.41}
\newcommand{\gsixVFullPuG}{0.120}
\newcommand{\gsixVFullPuH}{45.09}
\newcommand{\gsixVFullPuHOne}{18.41}
\newcommand{\gsixVFullSdH}{11.94}
\newcommand{\gsixVFullNrH}{14.53}
\newcommand{\gsixVFullNgH}{9.71}
\newcommand{\gsixVFullAllG}{1.976}
\newcommand{\gsixVFullAllH}{19.54}
\newcommand{\gsixVURatioFullIo}{4.8}
\newcommand{\gsixVURatioFullIoLo}{3.6}
\newcommand{\gsixVURatioFullIoHi}{6.6}
\newcommand{\gsixVHOneRatioFullIo}{7.5}
\newcommand{\gsixVHOneRatioFullIoLo}{5.2}
\newcommand{\gsixVHOneRatioFullIoHi}{11.8}
\newcommand{\gsixVURatioFullPu}{12.7}
\newcommand{\gsixVURatioFullPuLo}{4.1}
\newcommand{\gsixVURatioFullPuHi}{38.7}
\newcommand{\gsixVHOneRatioFullPu}{25.4}
\newcommand{\gsixVHOneRatioFullPuLo}{5.6}
\newcommand{\gsixVHOneRatioFullPuHi}{56.4}
\newcommand{\gsixVURatioFullSd}{10.1}
\newcommand{\gsixVURatioFullSdLo}{6.3}
\newcommand{\gsixVURatioFullSdHi}{19.5}
\newcommand{\gsixVHOneRatioFullSd}{11.9}
\newcommand{\gsixVHOneRatioFullSdLo}{6.4}
\newcommand{\gsixVHOneRatioFullSdHi}{29.6}
\newcommand{\gsixVURatioFullNr}{5.2}
\newcommand{\gsixVURatioFullNrLo}{1.9}
\newcommand{\gsixVURatioFullNrHi}{9.3}
\newcommand{\gsixVHOneRatioFullNr}{7.1}
\newcommand{\gsixVHOneRatioFullNrLo}{2.1}
\newcommand{\gsixVHOneRatioFullNrHi}{73.6}
\newcommand{\gsixVURatioFullNg}{8.9}
\newcommand{\gsixVURatioFullNgLo}{1.9}
\newcommand{\gsixVURatioFullNgHi}{38.7}
\newcommand{\gsixVHOneRatioFullNg}{11.6}
\newcommand{\gsixVHOneRatioFullNgLo}{3.0}
\newcommand{\gsixVHOneRatioFullNgHi}{43.6}
\newcommand{\gsixVURatioFullAll}{8.9}
\newcommand{\gsixVURatioFullAllLo}{5.2}
\newcommand{\gsixVURatioFullAllHi}{14.0}
\newcommand{\gsixVHOneRatioFullAll}{13.6}
\newcommand{\gsixVHOneRatioFullAllLo}{6.8}
\newcommand{\gsixVHOneRatioFullAllHi}{30.2}
\newcommand{\gsixVSkipShareIo}{6.1}
\newcommand{\gsixVSkipCatchSevereIo}{17.7}
\newcommand{\gsixVSkipSharePu}{50.4}
\newcommand{\gsixVSkipCatchSeverePu}{67.1}
\newcommand{\gsixVSkipShareSd}{2.0}
\newcommand{\gsixVSkipCatchSevereSd}{21.0}
\newcommand{\gsixVSkipShareNr}{7.6}
\newcommand{\gsixVSkipCatchSevereNr}{31.8}
\newcommand{\gsixVSkipShareNg}{4.8}
\newcommand{\gsixVSkipCatchSevereNg}{42.5}
\newcommand{\gsixVSkipShareAll}{6.8}
\newcommand{\gsixVSkipCatchSevereAll}{26.2}
\newcommand{\gsixVAucSevereMIo}{0.721}
\newcommand{\gsixVAucAnyMIo}{0.613}
\newcommand{\gsixVAucSevereMPu}{0.752}
\newcommand{\gsixVAucAnyMPu}{0.593}
\newcommand{\gsixVAucSevereMSd}{0.917}
\newcommand{\gsixVAucAnyMSd}{0.880}
\newcommand{\gsixVAucSevereMNr}{0.789}
\newcommand{\gsixVAucAnyMNr}{0.545}
\newcommand{\gsixVAucSevereMNg}{0.870}
\newcommand{\gsixVAucAnyMNg}{0.766}
\newcommand{\gsixVAucSevereMAll}{0.810}
\newcommand{\gsixVAucSevereMMin}{0.721}
\newcommand{\gsixVAucSevereMMax}{0.917}
\newcommand{\gsixVAucSevereMWins}{5}
\newcommand{\gsixVAucAnyMAll}{0.679}
\newcommand{\gsixVAucAnyMMin}{0.545}
\newcommand{\gsixVAucAnyMMax}{0.880}
\newcommand{\gsixVAucAnyMWins}{5}
\newcommand{\gsixVNDomPubAll}{6}
\newcommand{\gsixVExNames}{DNI U, DNI CH and Fisher CH}
\newcommand{\gsixVMoreGaindGMaxAll}{0.095}
\newcommand{\gsixVMoreGainURatioMinAll}{2.4}
\newcommand{\gsixVMoreGainSevRatioMinAll}{3.1}
\newcommand{\gsixVNMoreGainMainAll}{6}
\newcommand{\gsixVContGridNHarmed}{4{,}675}
\newcommand{\gsixVThK}{2}
\newcommand{\gsixVThKOne}{1}
\newcommand{\gsixVModelFeatures}{48}
\newcommand{\gsixVModelTrees}{400}
\newcommand{\gsixVModelLeafCap}{31}
\newcommand{\gsixVModelTrainCams}{4}
\newcommand{\gsixWGaussAllN}{153}
\newcommand{\gsixWGaussAllFullG}{3.564}
\newcommand{\gsixWGaussAllFullU}{0.1366}
\newcommand{\gsixWGaussAllFullH}{11.15}
\newcommand{\gsixWGaussAllFullHOne}{4.49}
\newcommand{\gsixWGaussAllTwoG}{3.506}
\newcommand{\gsixWGaussAllTwoU}{0.0117}
\newcommand{\gsixWGaussAllTwoH}{2.61}
\newcommand{\gsixWGaussAllTwoHOne}{0.29}
\newcommand{\gsixWGaussAllOneG}{3.176}
\newcommand{\gsixWGaussAllOneU}{0.0047}
\newcommand{\gsixWGaussAllOneH}{1.53}
\newcommand{\gsixWGaussAllOneHOne}{0.12}
\newcommand{\gsixWRealAllN}{278}
\newcommand{\gsixWRealAllFullG}{1.132}
\newcommand{\gsixWRealAllFullU}{0.2979}
\newcommand{\gsixWRealAllFullH}{23.76}
\newcommand{\gsixWRealAllFullHOne}{9.34}
\newcommand{\gsixWRealAllTwoG}{1.014}
\newcommand{\gsixWRealAllTwoU}{0.0291}
\newcommand{\gsixWRealAllTwoH}{8.70}
\newcommand{\gsixWRealAllTwoHOne}{0.57}
\newcommand{\gsixWRealAllOneG}{0.725}
\newcommand{\gsixWRealAllOneU}{0.0101}
\newcommand{\gsixWRealAllOneH}{6.10}
\newcommand{\gsixWRealAllOneHOne}{0.12}
\newcommand{\gsixWRegCOne}{$-$8.76}
\newcommand{\gsixWRegCOneLo}{$-$9.85}
\newcommand{\gsixWRegCOneHi}{$-$7.73}
\newcommand{\gsixWRegCTwo}{10.2}
\newcommand{\gsixWRegCTwoLo}{8.5}
\newcommand{\gsixWRegCTwoHi}{12.5}
\newcommand{\gsixWRegCThree}{$-$0.119}
\newcommand{\gsixWRegCThreeLo}{$-$0.174}
\newcommand{\gsixWRegCThreeHi}{$-$0.056}
\newcommand{\gsixBRealDniGG}{1.002}
\newcommand{\gsixBRealDniGU}{0.0951}
\newcommand{\gsixBRealDniGH}{15.68}
\newcommand{\gsixBRealDniGHOne}{2.59}
\newcommand{\gsixBRealDniUG}{1.096}
\newcommand{\gsixBRealDniUU}{0.1292}
\newcommand{\gsixBRealDniUH}{17.19}
\newcommand{\gsixBRealDniUHOne}{3.82}
\newcommand{\gsixBRealDniCHG}{1.184}
\newcommand{\gsixBRealDniCHU}{0.2448}
\newcommand{\gsixBRealDniCHH}{22.54}
\newcommand{\gsixBRealDniCHHOne}{7.95}
\newcommand{\gsixBRealSureCropG}{1.168}
\newcommand{\gsixBRealSureCropU}{0.1496}
\newcommand{\gsixBRealSureCropH}{17.77}
\newcommand{\gsixBRealSureCropHOne}{5.10}
\newcommand{\gsixBRealSureImgG}{1.158}
\newcommand{\gsixBRealSureImgU}{0.1666}
\newcommand{\gsixBRealSureImgH}{19.42}
\newcommand{\gsixBRealSureImgHOne}{5.59}
\newcommand{\gsixBRealSureUnitG}{1.194}
\newcommand{\gsixBRealSureUnitU}{0.1789}
\newcommand{\gsixBRealSureUnitH}{20.53}
\newcommand{\gsixBRealSureUnitHOne}{5.94}
\newcommand{\gsixBRealFisherCHG}{1.165}
\newcommand{\gsixBRealFisherCHU}{0.2587}
\newcommand{\gsixBRealFisherCHH}{22.94}
\newcommand{\gsixBRealFisherCHHOne}{8.37}
\newcommand{\gsixBRealPriorCapG}{0.853}
\newcommand{\gsixBRealPriorCapU}{0.1455}
\newcommand{\gsixBRealPriorCapH}{17.29}
\newcommand{\gsixBRealPriorCapHOne}{4.82}
\newcommand{\gsixBRealPriorHalfG}{0.891}
\newcommand{\gsixBRealPriorHalfU}{0.0908}
\newcommand{\gsixBRealPriorHalfH}{13.45}
\newcommand{\gsixBRealPriorHalfHOne}{2.34}
\newcommand{\gsixBRealCkdGG}{1.154}
\newcommand{\gsixBRealCkdGU}{0.0794}
\newcommand{\gsixBRealCkdGH}{13.53}
\newcommand{\gsixBRealCkdGHOne}{2.33}
\newcommand{\gsixBRealCkdUG}{1.222}
\newcommand{\gsixBRealCkdUU}{0.1359}
\newcommand{\gsixBRealCkdUH}{16.89}
\newcommand{\gsixBRealCkdUHOne}{4.13}
\newcommand{\gsixBGaussDniGG}{3.381}
\newcommand{\gsixBGaussDniGU}{0.0452}
\newcommand{\gsixBGaussDniGH}{6.77}
\newcommand{\gsixBGaussDniGHOne}{1.43}
\newcommand{\gsixBGaussDniUG}{3.525}
\newcommand{\gsixBGaussDniUU}{0.0597}
\newcommand{\gsixBGaussDniUH}{7.53}
\newcommand{\gsixBGaussDniUHOne}{2.01}
\newcommand{\gsixBGaussDniCHG}{3.624}
\newcommand{\gsixBGaussDniCHU}{0.1065}
\newcommand{\gsixBGaussDniCHH}{9.71}
\newcommand{\gsixBGaussDniCHHOne}{3.56}
\newcommand{\gsixBGaussSureCropG}{2.305}
\newcommand{\gsixBGaussSureCropU}{0.0163}
\newcommand{\gsixBGaussSureCropH}{2.10}
\newcommand{\gsixBGaussSureCropHOne}{0.52}
\newcommand{\gsixBGaussSureImgG}{2.356}
\newcommand{\gsixBGaussSureImgU}{0.0220}
\newcommand{\gsixBGaussSureImgH}{3.03}
\newcommand{\gsixBGaussSureImgHOne}{0.66}
\newcommand{\gsixBGaussSureUnitG}{2.394}
\newcommand{\gsixBGaussSureUnitU}{0.0288}
\newcommand{\gsixBGaussSureUnitH}{3.50}
\newcommand{\gsixBGaussSureUnitHOne}{0.91}
\newcommand{\gsixBGaussFisherCHG}{3.597}
\newcommand{\gsixBGaussFisherCHU}{0.1137}
\newcommand{\gsixBGaussFisherCHH}{10.10}
\newcommand{\gsixBGaussFisherCHHOne}{3.76}
\newcommand{\gsixBGaussPriorCapG}{1.923}
\newcommand{\gsixBGaussPriorCapU}{0.0600}
\newcommand{\gsixBGaussPriorCapH}{6.13}
\newcommand{\gsixBGaussPriorCapHOne}{2.14}
\newcommand{\gsixBGaussPriorHalfG}{2.277}
\newcommand{\gsixBGaussPriorHalfU}{0.0250}
\newcommand{\gsixBGaussPriorHalfH}{4.07}
\newcommand{\gsixBGaussPriorHalfHOne}{0.87}
\newcommand{\gsixBGaussCkdGG}{3.293}
\newcommand{\gsixBGaussCkdGU}{0.0153}
\newcommand{\gsixBGaussCkdGH}{2.82}
\newcommand{\gsixBGaussCkdGHOne}{0.45}
\newcommand{\gsixBGaussCkdUG}{3.569}
\newcommand{\gsixBGaussCkdUU}{0.0333}
\newcommand{\gsixBGaussCkdUH}{4.10}
\newcommand{\gsixBGaussCkdUHOne}{1.10}
\newcommand{\gsixBNPub}{9}
\newcommand{\gsixBNSettings}{5}
\newcommand{\gsixBPubNDom}{6}
\newcommand{\gsixBGateNDomWord}{one}
\newcommand{\gsixSrRealFullGHOne}{0.12}
\newcommand{\gsixSrRealDniGGHOne}{0.39}
\newcommand{\gsixSrRealDniUGHOne}{0.29}
\newcommand{\gsixSrRealDniCHGHOne}{0.15}
\newcommand{\gsixSrRealSureCropGHOne}{0.23}
\newcommand{\gsixSrRealSureImgGHOne}{0.21}
\newcommand{\gsixSrRealSureUnitGHOne}{0.20}
\newcommand{\gsixSrRealFisherCHGHOne}{0.14}
\newcommand{\gsixSrRealPriorCapGHOne}{0.18}
\newcommand{\gsixSrRealPriorHalfGHOne}{0.38}
\newcommand{\gsixSrRealCkdGGHOne}{0.50}
\newcommand{\gsixSrRealCkdUGHOne}{0.30}
\newcommand{\gsixSrRealOneGHOne}{6.02}
\newcommand{\gsixSrRealTwoGHOne}{1.76}
\newcommand{\gsixSrGaussFullGHOne}{0.79}
\newcommand{\gsixSrGaussDniGGHOne}{2.37}
\newcommand{\gsixSrGaussDniUGHOne}{1.76}
\newcommand{\gsixSrGaussDniCHGHOne}{1.02}
\newcommand{\gsixSrGaussSureCropGHOne}{4.46}
\newcommand{\gsixSrGaussSureImgGHOne}{3.55}
\newcommand{\gsixSrGaussSureUnitGHOne}{2.62}
\newcommand{\gsixSrGaussFisherCHGHOne}{0.96}
\newcommand{\gsixSrGaussPriorCapGHOne}{0.90}
\newcommand{\gsixSrGaussPriorHalfGHOne}{2.63}
\newcommand{\gsixSrGaussCkdGGHOne}{7.32}
\newcommand{\gsixSrGaussCkdUGHOne}{3.25}
\newcommand{\gsixSrGaussOneGHOne}{27.28}
\newcommand{\gsixSrGaussTwoGHOne}{12.06}
\newcommand{\hpAlphaPct}{5}
\newcommand{\hpCkList}{20, 50, 100, 200, 500 and 1000}
\newcommand{\gsixSkSpdGAll}{$-$0.095}
\newcommand{\gsixSkSpdGAllLo}{$-$0.137}
\newcommand{\gsixSkSpdGAllHi}{$-$0.060}
\newcommand{\gsixSkSpdHOneAll}{$-$0.29}
\newcommand{\gsixSkSpdHOneAllLo}{$-$0.43}
\newcommand{\gsixSkSpdHOneAllHi}{$-$0.16}
\newcommand{\gsixSkLodGAll}{$-$0.075}
\newcommand{\gsixSkLodGAllLo}{$-$0.113}
\newcommand{\gsixSkLodGAllHi}{$-$0.045}
\newcommand{\gsixSkLodHOneAll}{$-$0.23}
\newcommand{\gsixSkLodHOneAllLo}{$-$0.35}
\newcommand{\gsixSkLodHOneAllHi}{$-$0.12}
\newcommand{\gsixSkLoFullIoG}{1.664}
\newcommand{\gsixSkLoFullIoHOne}{5.44}
\newcommand{\gsixSkLoFullPuG}{0.120}
\newcommand{\gsixSkLoFullPuHOne}{18.41}
\newcommand{\gsixSkLoFullSdG}{4.617}
\newcommand{\gsixSkLoFullSdHOne}{5.40}
\newcommand{\gsixSkLoFullNrG}{0.916}
\newcommand{\gsixSkLoFullNrHOne}{3.02}
\newcommand{\gsixSkLoFullNgG}{2.562}
\newcommand{\gsixSkLoFullNgHOne}{3.56}
\newcommand{\gsixSkLoFullAllG}{1.976}
\newcommand{\gsixSkLoFullAllHOne}{7.17}
\newcommand{\gsixSkLoMeanIoG}{1.535}
\newcommand{\gsixSkLoMeanIoHOne}{1.96}
\newcommand{\gsixSkLoMeanPuG}{0.327}
\newcommand{\gsixSkLoMeanPuHOne}{4.56}
\newcommand{\gsixSkLoMeanSdG}{4.754}
\newcommand{\gsixSkLoMeanSdHOne}{0.89}
\newcommand{\gsixSkLoMeanNrG}{0.879}
\newcommand{\gsixSkLoMeanNrHOne}{0.97}
\newcommand{\gsixSkLoMeanNgG}{2.491}
\newcommand{\gsixSkLoMeanNgHOne}{0.51}
\newcommand{\gsixSkLoMeanAllG}{1.997}
\newcommand{\gsixSkLoMeanAllHOne}{1.78}
\newcommand{\gsixSkLoQuantIoG}{1.533}
\newcommand{\gsixSkLoQuantIoHOne}{1.02}
\newcommand{\gsixSkLoQuantPuG}{0.285}
\newcommand{\gsixSkLoQuantPuHOne}{0.82}
\newcommand{\gsixSkLoQuantSdG}{4.795}
\newcommand{\gsixSkLoQuantSdHOne}{0.75}
\newcommand{\gsixSkLoQuantNrG}{0.819}
\newcommand{\gsixSkLoQuantNrHOne}{0.57}
\newcommand{\gsixSkLoQuantNgG}{2.559}
\newcommand{\gsixSkLoQuantNgHOne}{0.62}
\newcommand{\gsixSkLoQuantAllG}{1.998}
\newcommand{\gsixSkLoQuantAllHOne}{0.76}
\newcommand{\gsixSkLoTDIoG}{1.373}
\newcommand{\gsixSkLoTDIoHOne}{0.72}
\newcommand{\gsixSkLoTDPuG}{0.265}
\newcommand{\gsixSkLoTDPuHOne}{0.72}
\newcommand{\gsixSkLoTDSdG}{4.738}
\newcommand{\gsixSkLoTDSdHOne}{0.45}
\newcommand{\gsixSkLoTDNrG}{0.795}
\newcommand{\gsixSkLoTDNrHOne}{0.43}
\newcommand{\gsixSkLoTDNgG}{2.444}
\newcommand{\gsixSkLoTDNgHOne}{0.31}
\newcommand{\gsixSkLoTDAllG}{1.923}
\newcommand{\gsixSkLoTDAllHOne}{0.53}
\newcommand{\gsixSkLoHalfIoG}{1.176}
\newcommand{\gsixSkLoHalfIoHOne}{0.78}
\newcommand{\gsixSkLoHalfPuG}{0.228}
\newcommand{\gsixSkLoHalfPuHOne}{6.24}
\newcommand{\gsixSkLoHalfSdG}{2.824}
\newcommand{\gsixSkLoHalfSdHOne}{1.20}
\newcommand{\gsixSkLoHalfNrG}{0.713}
\newcommand{\gsixSkLoHalfNrHOne}{0.49}
\newcommand{\gsixSkLoHalfNgG}{1.775}
\newcommand{\gsixSkLoHalfNgHOne}{0.60}
\newcommand{\gsixSkLoHalfAllG}{1.343}
\newcommand{\gsixSkLoHalfAllHOne}{1.86}
\newcommand{\gsixSkLoCapIoG}{1.170}
\newcommand{\gsixSkLoCapIoHOne}{2.42}
\newcommand{\gsixSkLoCapPuG}{0.183}
\newcommand{\gsixSkLoCapPuHOne}{10.99}
\newcommand{\gsixSkLoCapSdG}{2.071}
\newcommand{\gsixSkLoCapSdHOne}{2.21}
\newcommand{\gsixSkLoCapNrG}{0.691}
\newcommand{\gsixSkLoCapNrHOne}{1.80}
\newcommand{\gsixSkLoCapNgG}{1.684}
\newcommand{\gsixSkLoCapNgHOne}{1.97}
\newcommand{\gsixSkLoCapAllG}{1.159}
\newcommand{\gsixSkLoCapAllHOne}{3.88}
\newcommand{\gsixSkSpFullIoG}{1.495}
\newcommand{\gsixSkSpFullIoHOne}{7.09}
\newcommand{\gsixSkSpFullPuG}{-0.263}
\newcommand{\gsixSkSpFullPuHOne}{27.32}
\newcommand{\gsixSkSpFullSdG}{4.524}
\newcommand{\gsixSkSpFullSdHOne}{5.26}
\newcommand{\gsixSkSpFullNrG}{0.862}
\newcommand{\gsixSkSpFullNrHOne}{3.24}
\newcommand{\gsixSkSpFullNgG}{2.484}
\newcommand{\gsixSkSpFullNgHOne}{3.62}
\newcommand{\gsixSkSpFullAllG}{1.820}
\newcommand{\gsixSkSpFullAllHOne}{9.31}
\newcommand{\gsixSkSpMeanIoG}{1.406}
\newcommand{\gsixSkSpMeanIoHOne}{2.18}
\newcommand{\gsixSkSpMeanPuG}{0.282}
\newcommand{\gsixSkSpMeanPuHOne}{4.91}
\newcommand{\gsixSkSpMeanSdG}{4.583}
\newcommand{\gsixSkSpMeanSdHOne}{0.65}
\newcommand{\gsixSkSpMeanNrG}{0.830}
\newcommand{\gsixSkSpMeanNrHOne}{0.78}
\newcommand{\gsixSkSpMeanNgG}{2.385}
\newcommand{\gsixSkSpMeanNgHOne}{0.24}
\newcommand{\gsixSkSpMeanAllG}{1.897}
\newcommand{\gsixSkSpMeanAllHOne}{1.75}
\newcommand{\gsixSkSpQuantIoG}{1.396}
\newcommand{\gsixSkSpQuantIoHOne}{1.08}
\newcommand{\gsixSkSpQuantPuG}{0.276}
\newcommand{\gsixSkSpQuantPuHOne}{0.44}
\newcommand{\gsixSkSpQuantSdG}{4.673}
\newcommand{\gsixSkSpQuantSdHOne}{0.68}
\newcommand{\gsixSkSpQuantNrG}{0.766}
\newcommand{\gsixSkSpQuantNrHOne}{0.74}
\newcommand{\gsixSkSpQuantNgG}{2.487}
\newcommand{\gsixSkSpQuantNgHOne}{0.47}
\newcommand{\gsixSkSpQuantAllG}{1.920}
\newcommand{\gsixSkSpQuantAllHOne}{0.68}
\newcommand{\gsixSkSpTDIoG}{1.241}
\newcommand{\gsixSkSpTDIoHOne}{0.68}
\newcommand{\gsixSkSpTDPuG}{0.258}
\newcommand{\gsixSkSpTDPuHOne}{0.26}
\newcommand{\gsixSkSpTDSdG}{4.546}
\newcommand{\gsixSkSpTDSdHOne}{0.36}
\newcommand{\gsixSkSpTDNrG}{0.744}
\newcommand{\gsixSkSpTDNrHOne}{0.43}
\newcommand{\gsixSkSpTDNgG}{2.337}
\newcommand{\gsixSkSpTDNgHOne}{0.21}
\newcommand{\gsixSkSpTDAllG}{1.825}
\newcommand{\gsixSkSpTDAllHOne}{0.39}
\newcommand{\gsixSkSpHalfIoG}{1.102}
\newcommand{\gsixSkSpHalfIoHOne}{1.21}
\newcommand{\gsixSkSpHalfPuG}{0.119}
\newcommand{\gsixSkSpHalfPuHOne}{9.99}
\newcommand{\gsixSkSpHalfSdG}{2.728}
\newcommand{\gsixSkSpHalfSdHOne}{0.95}
\newcommand{\gsixSkSpHalfNrG}{0.701}
\newcommand{\gsixSkSpHalfNrHOne}{0.47}
\newcommand{\gsixSkSpHalfNgG}{1.770}
\newcommand{\gsixSkSpHalfNgHOne}{0.78}
\newcommand{\gsixSkSpHalfAllG}{1.284}
\newcommand{\gsixSkSpHalfAllHOne}{2.68}
\newcommand{\gsixSkSpCapIoG}{1.063}
\newcommand{\gsixSkSpCapIoHOne}{3.53}
\newcommand{\gsixSkSpCapPuG}{0.080}
\newcommand{\gsixSkSpCapPuHOne}{14.34}
\newcommand{\gsixSkSpCapSdG}{2.147}
\newcommand{\gsixSkSpCapSdHOne}{2.37}
\newcommand{\gsixSkSpCapNrG}{0.667}
\newcommand{\gsixSkSpCapNrHOne}{1.99}
\newcommand{\gsixSkSpCapNgG}{1.670}
\newcommand{\gsixSkSpCapNgHOne}{1.88}
\newcommand{\gsixSkSpCapAllG}{1.126}
\newcommand{\gsixSkSpCapAllHOne}{4.82}
\newcommand{\gsixSkLoFullAllGH}{0.101}
\newcommand{\gsixSkLoFullAllGHOne}{0.28}
\newcommand{\gsixSkLoMeanAllGH}{0.143}
\newcommand{\gsixSkLoMeanAllGHOne}{1.12}
\newcommand{\gsixSkLoQuantAllGH}{0.241}
\newcommand{\gsixSkLoQuantAllGHOne}{2.64}
\newcommand{\gsixSkLoTDAllGH}{0.254}
\newcommand{\gsixSkLoTDAllGHOne}{3.65}
\newcommand{\gsixSkLoHalfAllGH}{0.113}
\newcommand{\gsixSkLoHalfAllGHOne}{0.72}
\newcommand{\gsixSkLoCapAllGH}{0.081}
\newcommand{\gsixSkLoCapAllGHOne}{0.30}
\newcommand{\gsixSkSpFullAllGH}{0.081}
\newcommand{\gsixSkSpFullAllGHOne}{0.20}
\newcommand{\gsixSkSpMeanAllGH}{0.132}
\newcommand{\gsixSkSpMeanAllGHOne}{1.08}
\newcommand{\gsixSkSpQuantAllGH}{0.245}
\newcommand{\gsixSkSpQuantAllGHOne}{2.82}
\newcommand{\gsixSkSpTDAllGH}{0.256}
\newcommand{\gsixSkSpTDAllGHOne}{4.71}
\newcommand{\gsixSkSpHalfAllGH}{0.097}
\newcommand{\gsixSkSpHalfAllGHOne}{0.48}
\newcommand{\gsixSkSpCapAllGH}{0.070}
\newcommand{\gsixSkSpCapAllGHOne}{0.23}
\newcommand{\gsixTsOneMCTwo}{8.9}
\newcommand{\gsixTsTwoMCTwo}{10.2}
\newcommand{\gsixTsOneN}{72}
\newcommand{\gsixTsTwoN}{278}
\newcommand{\gsixParNCameras}{20}
\newcommand{\gsixParNSources}{5}
\newcommand{\gsixParLowN}{431}
\newcommand{\gsixParLowFullHOne}{7.17}
\newcommand{\gsixParLowTDHOne}{0.53}
\newcommand{\gsixParLowRatio}{13.6}
\newcommand{\gsixParLowdG}{$-$0.05}
\newcommand{\gsixParSpN}{431}
\newcommand{\gsixParSpFullHOne}{9.31}
\newcommand{\gsixParSpTDHOne}{0.39}
\newcommand{\gsixParSpFullG}{1.820}
\newcommand{\gsixParSpURatio}{14.4}
\newcommand{\gsixParSpTDG}{1.825}
\newcommand{\gsixParSpRatio}{24.0}
\newcommand{\gsixParSpdG}{+0.01}
\newcommand{\gsixParSpRatioLo}{15.4}
\newcommand{\gsixParSpdGLo}{$-$0.11}
\newcommand{\gsixParSpRatioHi}{44.1}
\newcommand{\gsixParSpdGHi}{+0.14}
\newcommand{\gsixYNCompRows}{11}
\newcommand{\gsixYCamPu}{2}
\newcommand{\gsixYCamSd}{3}
\newcommand{\gsixYCamAll}{20}
\newcommand{\gsixSpReNUnits}{278}
\newcommand{\gsixSpReFullG}{0.900}
\newcommand{\gsixSpReFullU}{0.2909}
\newcommand{\gsixSpReFullH}{25.34}
\newcommand{\gsixSpReFullHOne}{8.95}
\newcommand{\gsixSpReURatioFull}{8.7 [4.4, 14.8]}
\newcommand{\gsixSpReSevRatioFull}{14.3 [6.0, 42.0]}
\newcommand{\gsixSpReDniGG}{0.848}
\newcommand{\gsixSpReDniGU}{0.0948}
\newcommand{\gsixSpReDniGH}{18.45}
\newcommand{\gsixSpReDniGHOne}{2.64}
\newcommand{\gsixSpReURatioDniG}{2.8 [1.5, 5.4]}
\newcommand{\gsixSpReSevRatioDniG}{4.2 [1.7, 12.6]}
\newcommand{\gsixSpReDniUG}{0.910}
\newcommand{\gsixSpReDniUU}{0.1300}
\newcommand{\gsixSpReDniUH}{19.80}
\newcommand{\gsixSpReDniUHOne}{3.99}
\newcommand{\gsixSpReURatioDniU}{3.9 [2.0, 7.1]}
\newcommand{\gsixSpReSevRatioDniU}{6.4 [2.6, 18.8]}
\newcommand{\gsixSpReDniCHG}{0.937}
\newcommand{\gsixSpReDniCHU}{0.2466}
\newcommand{\gsixSpReDniCHH}{24.13}
\newcommand{\gsixSpReDniCHHOne}{7.77}
\newcommand{\gsixSpReURatioDniCH}{7.4 [4.0, 12.6]}
\newcommand{\gsixSpReSevRatioDniCH}{12.5 [5.4, 35.9]}
\newcommand{\gsixSpReSureCropG}{0.921}
\newcommand{\gsixSpReSureCropU}{0.1578}
\newcommand{\gsixSpReSureCropH}{20.81}
\newcommand{\gsixSpReSureCropHOne}{5.37}
\newcommand{\gsixSpReURatioSureCrop}{4.7 [2.4, 9.0]}
\newcommand{\gsixSpReSevRatioSureCrop}{8.6 [3.5, 26.3]}
\newcommand{\gsixSpReSureImgG}{0.928}
\newcommand{\gsixSpReSureImgU}{0.1727}
\newcommand{\gsixSpReSureImgH}{22.30}
\newcommand{\gsixSpReSureImgHOne}{5.88}
\newcommand{\gsixSpReURatioSureImg}{5.2 [2.8, 9.9]}
\newcommand{\gsixSpReSevRatioSureImg}{9.4 [3.9, 28.4]}
\newcommand{\gsixSpReSureUnitG}{0.967}
\newcommand{\gsixSpReSureUnitU}{0.1799}
\newcommand{\gsixSpReSureUnitH}{22.96}
\newcommand{\gsixSpReSureUnitHOne}{6.32}
\newcommand{\gsixSpReURatioSureUnit}{5.4 [2.9, 10.2]}
\newcommand{\gsixSpReSevRatioSureUnit}{10.1 [4.0, 31.3]}
\newcommand{\gsixSpReFisherCHG}{0.914}
\newcommand{\gsixSpReFisherCHU}{0.2685}
\newcommand{\gsixSpReFisherCHH}{24.59}
\newcommand{\gsixSpReFisherCHHOne}{8.06}
\newcommand{\gsixSpReURatioFisherCH}{8.0 [4.2, 13.5]}
\newcommand{\gsixSpReSevRatioFisherCH}{12.9 [5.6, 37.4]}
\newcommand{\gsixSpRePriorCapG}{0.681}
\newcommand{\gsixSpRePriorCapU}{0.1491}
\newcommand{\gsixSpRePriorCapH}{19.69}
\newcommand{\gsixSpRePriorCapHOne}{5.07}
\newcommand{\gsixSpReURatioPriorCap}{4.5 [1.8, 9.4]}
\newcommand{\gsixSpReSevRatioPriorCap}{8.1 [2.5, 27.4]}
\newcommand{\gsixSpRePriorHalfG}{0.705}
\newcommand{\gsixSpRePriorHalfU}{0.1072}
\newcommand{\gsixSpRePriorHalfH}{16.82}
\newcommand{\gsixSpRePriorHalfHOne}{2.50}
\newcommand{\gsixSpReURatioPriorHalf}{3.2 [1.2, 5.7]}
\newcommand{\gsixSpReSevRatioPriorHalf}{4.0 [0.97, 13.2]}
\newcommand{\gsixSpReCkdGG}{0.854}
\newcommand{\gsixSpReCkdGU}{0.0786}
\newcommand{\gsixSpReCkdGH}{16.47}
\newcommand{\gsixSpReCkdGHOne}{2.07}
\newcommand{\gsixSpReURatioCkdG}{2.3 [1.01, 4.3]}
\newcommand{\gsixSpReSevRatioCkdG}{3.3 [1.04, 10.3]}
\newcommand{\gsixSpReCkdUG}{0.938}
\newcommand{\gsixSpReCkdUU}{0.1389}
\newcommand{\gsixSpReCkdUH}{19.29}
\newcommand{\gsixSpReCkdUHOne}{3.97}
\newcommand{\gsixSpReURatioCkdU}{4.2 [1.9, 7.4]}
\newcommand{\gsixSpReSevRatioCkdU}{6.4 [2.1, 20.2]}
\newcommand{\gsixSpReMeanMG}{0.914}
\newcommand{\gsixSpReMeanMU}{0.1026}
\newcommand{\gsixSpReMeanMH}{20.10}
\newcommand{\gsixSpReMeanMHOne}{2.50}
\newcommand{\gsixSpReURatioMeanM}{3.1 [2.0, 5.0]}
\newcommand{\gsixSpReSevRatioMeanM}{4.0 [2.4, 8.8]}
\newcommand{\gsixSpReQOnlyG}{0.879}
\newcommand{\gsixSpReQOnlyU}{0.0382}
\newcommand{\gsixSpReQOnlyH}{10.83}
\newcommand{\gsixSpReQOnlyHOne}{0.80}
\newcommand{\gsixSpReURatioQOnly}{1.1 [1.1, 1.2]}
\newcommand{\gsixSpReSevRatioQOnly}{1.3 [1.1, 1.6]}
\newcommand{\gsixSpReKOneG}{0.585}
\newcommand{\gsixSpReKOneU}{0.0122}
\newcommand{\gsixSpReKOneH}{7.69}
\newcommand{\gsixSpReKOneHOne}{0.19}
\newcommand{\gsixSpReURatioKOne}{0.36 [0.33, 0.40]}
\newcommand{\gsixSpReSevRatioKOne}{0.30 [0.14, 0.45]}
\newcommand{\gsixSpReKTwoG}{0.811}
\newcommand{\gsixSpReKTwoU}{0.0335}
\newcommand{\gsixSpReKTwoH}{10.42}
\newcommand{\gsixSpReKTwoHOne}{0.62}
\newcommand{\gsixSpGaNUnits}{153}
\newcommand{\gsixSpGaFullG}{3.590}
\newcommand{\gsixSpGaFullU}{0.1431}
\newcommand{\gsixSpGaFullH}{10.83}
\newcommand{\gsixSpGaFullHOne}{4.48}
\newcommand{\gsixSpGaURatioFull}{9.7 [5.8, 17.9]}
\newcommand{\gsixSpGaSevRatioFull}{11.8 [6.6, 23.6]}
\newcommand{\gsixSpGaDniGG}{3.454}
\newcommand{\gsixSpGaDniGU}{0.0488}
\newcommand{\gsixSpGaDniGH}{6.68}
\newcommand{\gsixSpGaDniGHOne}{1.40}
\newcommand{\gsixSpGaURatioDniG}{3.3 [2.0, 5.9]}
\newcommand{\gsixSpGaSevRatioDniG}{3.7 [2.0, 6.9]}
\newcommand{\gsixSpGaDniUG}{3.590}
\newcommand{\gsixSpGaDniUU}{0.0638}
\newcommand{\gsixSpGaDniUH}{7.29}
\newcommand{\gsixSpGaDniUHOne}{1.84}
\newcommand{\gsixSpGaURatioDniU}{4.3 [2.6, 7.8]}
\newcommand{\gsixSpGaSevRatioDniU}{4.9 [2.6, 9.3]}
\newcommand{\gsixSpGaDniCHG}{3.638}
\newcommand{\gsixSpGaDniCHU}{0.1238}
\newcommand{\gsixSpGaDniCHH}{9.85}
\newcommand{\gsixSpGaDniCHHOne}{3.90}
\newcommand{\gsixSpGaURatioDniCH}{8.4 [4.9, 15.9]}
\newcommand{\gsixSpGaSevRatioDniCH}{10.3 [5.7, 20.5]}
\newcommand{\gsixSpGaSureCropG}{2.413}
\newcommand{\gsixSpGaSureCropU}{0.0172}
\newcommand{\gsixSpGaSureCropH}{2.40}
\newcommand{\gsixSpGaSureCropHOne}{0.48}
\newcommand{\gsixSpGaURatioSureCrop}{1.2 [0.49, 2.6]}
\newcommand{\gsixSpGaSevRatioSureCrop}{1.3 [0.52, 2.7]}
\newcommand{\gsixSpGaSureImgG}{2.475}
\newcommand{\gsixSpGaSureImgU}{0.0231}
\newcommand{\gsixSpGaSureImgH}{3.24}
\newcommand{\gsixSpGaSureImgHOne}{0.67}
\newcommand{\gsixSpGaURatioSureImg}{1.6 [0.72, 3.2]}
\newcommand{\gsixSpGaSevRatioSureImg}{1.8 [0.68, 4.1]}
\newcommand{\gsixSpGaSureUnitG}{2.534}
\newcommand{\gsixSpGaSureUnitU}{0.0312}
\newcommand{\gsixSpGaSureUnitH}{3.80}
\newcommand{\gsixSpGaSureUnitHOne}{0.95}
\newcommand{\gsixSpGaURatioSureUnit}{2.1 [0.91, 4.7]}
\newcommand{\gsixSpGaSevRatioSureUnit}{2.5 [0.96, 6.0]}
\newcommand{\gsixSpGaFisherCHG}{3.611}
\newcommand{\gsixSpGaFisherCHU}{0.1300}
\newcommand{\gsixSpGaFisherCHH}{10.13}
\newcommand{\gsixSpGaFisherCHHOne}{4.11}
\newcommand{\gsixSpGaURatioFisherCH}{8.8 [5.3, 16.6]}
\newcommand{\gsixSpGaSevRatioFisherCH}{10.8 [6.1, 21.6]}
\newcommand{\gsixSpGaPriorCapG}{1.877}
\newcommand{\gsixSpGaPriorCapU}{0.0608}
\newcommand{\gsixSpGaPriorCapH}{6.41}
\newcommand{\gsixSpGaPriorCapHOne}{2.09}
\newcommand{\gsixSpGaURatioPriorCap}{4.1 [2.4, 7.5]}
\newcommand{\gsixSpGaSevRatioPriorCap}{5.5 [2.5, 12.3]}
\newcommand{\gsixSpGaPriorHalfG}{2.299}
\newcommand{\gsixSpGaPriorHalfU}{0.0293}
\newcommand{\gsixSpGaPriorHalfH}{4.48}
\newcommand{\gsixSpGaPriorHalfHOne}{0.90}
\newcommand{\gsixSpGaURatioPriorHalf}{2.0 [1.1, 3.7]}
\newcommand{\gsixSpGaSevRatioPriorHalf}{2.4 [1.01, 5.0]}
\newcommand{\gsixSpGaCkdGG}{3.562}
\newcommand{\gsixSpGaCkdGU}{0.0366}
\newcommand{\gsixSpGaCkdGH}{4.47}
\newcommand{\gsixSpGaCkdGHOne}{1.04}
\newcommand{\gsixSpGaURatioCkdG}{2.5 [1.4, 4.8]}
\newcommand{\gsixSpGaSevRatioCkdG}{2.7 [1.4, 5.4]}
\newcommand{\gsixSpGaCkdUG}{3.535}
\newcommand{\gsixSpGaCkdUU}{0.0335}
\newcommand{\gsixSpGaCkdUH}{4.23}
\newcommand{\gsixSpGaCkdUHOne}{0.96}
\newcommand{\gsixSpGaURatioCkdU}{2.3 [1.2, 4.4]}
\newcommand{\gsixSpGaSevRatioCkdU}{2.5 [1.2, 5.1]}
\newcommand{\gsixSpGaMeanMG}{3.623}
\newcommand{\gsixSpGaMeanMU}{0.0241}
\newcommand{\gsixSpGaMeanMH}{4.72}
\newcommand{\gsixSpGaMeanMHOne}{0.70}
\newcommand{\gsixSpGaURatioMeanM}{1.6 [1.2, 2.2]}
\newcommand{\gsixSpGaSevRatioMeanM}{1.8 [1.2, 3.1]}
\newcommand{\gsixSpGaQOnlyG}{3.677}
\newcommand{\gsixSpGaQOnlyU}{0.0247}
\newcommand{\gsixSpGaQOnlyH}{4.47}
\newcommand{\gsixSpGaQOnlyHOne}{0.69}
\newcommand{\gsixSpGaURatioQOnly}{1.7 [1.4, 2.0]}
\newcommand{\gsixSpGaSevRatioQOnly}{1.8 [1.3, 2.3]}
\newcommand{\gsixSpGaKOneG}{3.280}
\newcommand{\gsixSpGaKOneU}{0.0057}
\newcommand{\gsixSpGaKOneH}{1.99}
\newcommand{\gsixSpGaKOneHOne}{0.09}
\newcommand{\gsixSpGaURatioKOne}{0.38 [0.28, 0.46]}
\newcommand{\gsixSpGaSevRatioKOne}{0.25 [0.03, 0.38]}
\newcommand{\gsixSpGaKTwoG}{3.591}
\newcommand{\gsixSpGaKTwoU}{0.0148}
\newcommand{\gsixSpGaKTwoH}{3.32}
\newcommand{\gsixSpGaKTwoHOne}{0.38}
\newcommand{\gsixSpReFullGHOne}{0.10}
\newcommand{\gsixSpReDniGGHOne}{0.32}
\newcommand{\gsixSpReDniUGHOne}{0.23}
\newcommand{\gsixSpReDniCHGHOne}{0.12}
\newcommand{\gsixSpReSureCropGHOne}{0.17}
\newcommand{\gsixSpReSureImgGHOne}{0.16}
\newcommand{\gsixSpReSureUnitGHOne}{0.15}
\newcommand{\gsixSpReFisherCHGHOne}{0.11}
\newcommand{\gsixSpRePriorCapGHOne}{0.13}
\newcommand{\gsixSpRePriorHalfGHOne}{0.28}
\newcommand{\gsixSpReCkdGGHOne}{0.41}
\newcommand{\gsixSpReCkdUGHOne}{0.24}
\newcommand{\gsixSpReMeanMGHOne}{0.37}
\newcommand{\gsixSpReQOnlyGHOne}{1.10}
\newcommand{\gsixSpReKOneGHOne}{3.10}
\newcommand{\gsixSpReKTwoGHOne}{1.30}
\newcommand{\gsixSpGaFullGHOne}{0.80}
\newcommand{\gsixSpGaDniGGHOne}{2.47}
\newcommand{\gsixSpGaDniUGHOne}{1.95}
\newcommand{\gsixSpGaDniCHGHOne}{0.93}
\newcommand{\gsixSpGaSureCropGHOne}{5.06}
\newcommand{\gsixSpGaSureImgGHOne}{3.69}
\newcommand{\gsixSpGaSureUnitGHOne}{2.67}
\newcommand{\gsixSpGaFisherCHGHOne}{0.88}
\newcommand{\gsixSpGaPriorCapGHOne}{0.90}
\newcommand{\gsixSpGaPriorHalfGHOne}{2.55}
\newcommand{\gsixSpGaCkdGGHOne}{3.43}
\newcommand{\gsixSpGaCkdUGHOne}{3.68}
\newcommand{\gsixSpGaMeanMGHOne}{5.17}
\newcommand{\gsixSpGaQOnlyGHOne}{5.36}
\newcommand{\gsixSpGaKOneGHOne}{34.93}
\newcommand{\gsixSpGaKTwoGHOne}{9.45}
\newcommand{\gsixSpReGHOneBestComp}{0.41}
\newcommand{\gsixSpGaGHOneBestComp}{5.06}
\newcommand{\gsixSpReGainOverFull}{$-$0.089}
\newcommand{\gsixSpReGainOverFullLo}{$-$0.189}
\newcommand{\gsixSpReGainOverFullHi}{+0.047}
\newcommand{\gsixSpReMoreGainURatioMin}{2.3}
\newcommand{\gsixSpGaGainOverFull}{+0.001}
\newcommand{\gsixSpGaGainOverFullLo}{$-$0.078}
\newcommand{\gsixSpGaGainOverFullHi}{+0.086}
\newcommand{\gsixSpGaMoreGainURatioMin}{8.4}
\newcommand{\gsixSpBlockRe}{Real-noise-trained sources (IOCI, PolyU, NIND)}
\newcommand{\gsixSpBlockGa}{Gaussian-trained sources (SIDD, NIND)}
\newcommand{\gsixSpCaption}{\textbf{No published controller, label-free gate or full adaptation dominates \ours{} in gain and downside, with either kind of source network.} Each block weights its settings equally (\gsixSpReNUnits{} and \gsixSpGaNUnits{} low-rank units), ratios divide each controller's downside or severe-harm share by that of \ours{} with camera-clustered \gsixBootLevel{} percent intervals (\gsixBootN{} replicates), and bold marks the best value per column outside the $k=\gsixVThKOne{}$ row. \ours{} has the lowest downside of every published controller, gate and full adaptation in both blocks, and the lowest harmed and severe-harm shares with real-noise-trained sources. Every published controller, gate row or full adaptation with more gain has at least \gsixSpReMoreGainURatioMin{} (real-noise) and \gsixSpGaMoreGainURatioMin{} (Gaussian) times the downside of \ours{}. $H$ and $H_1$ are the harmed and severe-harm shares (\%), and $G/H_1$ is the gain per severely harmed percentage point of \cref{sec:problem}, highest for the harm-averse \ours{} ($k=\gsixVThKOne{}$), as expected of a cautious rule.}
\newcommand{\suppPerSrcTab}{S3}

\newcommand{\ours}{TailDose}
\newcommand{\PropRef}[1]{\cref{prop:#1}}
\newcommand{\PropHarmTitle}{threshold, monotonicity and containment}
\newcommand{\PropHarmBody}{%
Assume $E,R_0>0$ and $f_{\dep}(x)\neq y$, and fix a crop and $c\in[0,1]$. In words: harm needs an alignment below half the dose, so a dose of at most twice a forecast confines harm to forecast misses.
\begin{enumerate}[label=(\roman*),leftmargin=*,nosep,topsep=2pt]
\item \textbf{Threshold.} The crop is harmed iff $c>0$ and $t<c/2$.
\item \textbf{Monotone in the dose.} The squared-error downside $u_c=(-D_c)_+=E\,c\,(c-2t)_+$ is nondecreasing in $c$. A crop harmed at $c$ is harmed by full adaptation, by no less in dB.
\item \textbf{Containment.} For any score $\hat q$ and $c\le\max(0,2\hat q)$, harm implies $t<\hat q$, so the harmed share is at most the miss share $\Pr(t<\hat q)$ for any weighting of crops.
\end{enumerate}}
\newcommand{\PropCCTitle}{chance-constrained dose}
\newcommand{\PropCCBody}{%
Fix $\phi$ with $E>0$ and let the conditional law of $t$ given $\phi$ have no atoms, a finite mean $\tb=\E[t\mid\phi]$ and distribution function $F$. Let $\alpha\in(0,1)$ and $q_\alpha$ be the largest $\alpha$-quantile of $F$. In words: under a cap $\alpha$ on the harm probability, the best dose is twice the $\alpha$-quantile of the alignment, capped at its mean.
Among $c\in[0,1]$ with $\Pr(\Delta_c<0\mid\phi)\le\alpha$, the expected reduction in squared error $\E[D_c\mid\phi]=E\,c\,(2\tb-c)$ is maximised by $c^\star_\alpha=\clip(\min(\tb,2q_\alpha),0,1)$.}
\newcommand{\PropSuffTitle}{realised level}
\newcommand{\PropSuffBody}{%
Let $\hat q$ be a fixed function of $\phi$, let the conditional distribution function $F(\cdot\mid\phi)$ of $t$ on the test camera be continuous and strictly increasing, and call $a(\phi)=F(\hat q(\phi)\mid\phi)$ the \emph{realised level}. In words: on a new camera the forecast is a quantile at an unknown level, which is its miss rate and bounds the harm probability of \ours{} on crops with a dose strictly between zero and one.
\begin{enumerate}[label=(\roman*),leftmargin=*,nosep,topsep=2pt]
\item \textbf{Miss rate.} $\hat q=q_{a(\phi)}$, and the marginal miss rate is $\Pr(t<\hat q)=\E[a(\phi)]$.
\item \textbf{Harm under the dose.} Let $\hat\mu$ be a fixed function of $\phi$, $0<k\le2$ and $c=\clip(\min(\hat\mu,k\hat q),0,1)$ with $0<c<1$. The harm probability of a crop given $\phi$ is $F(c/2\mid\phi)\le F(k\hat q/2\mid\phi)\le a(\phi)$, with equality in the first where $c=k\hat q$ and in the second at $k=2$. At $k=2$ and $\hat\mu=\tb$, $c$ is the target of \PropRef{cc} at level $a(\phi)$.
\end{enumerate}}
\let\gsixSpCaptionGen\gsixSpCaption
\renewcommand{\gsixSpCaption}{\gsixSpCaptionGen{} \ours{} minus full adaptation in gain is \gsixSpReGainOverFull{}~dB [\gsixSpReGainOverFullLo{}, \gsixSpReGainOverFullHi{}] with real-noise-trained and \gsixSpGaGainOverFull{}~dB [\gsixSpGaGainOverFullLo{}, \gsixSpGaGainOverFullHi{}] with Gaussian-trained sources. DNI G, DNI U and the gate were matched on development cameras to the gain (G) or downside (U) of an earlier quantile rule, not to \ours{}. The values pooled over the settings and $G/H$ per block are in the supplementary material.}
\newcommand{\gsixMMissIoTwo}{9.95}
\newcommand{\gsixMMissPu}{30.8}
\newcommand{\gsixMMissPuTwo}{30.76}
\newcommand{\gsixMMissSdTwo}{9.33}
\newcommand{\gsixMMissNrTwo}{10.04}
\newcommand{\gsixMMissNgTwo}{7.03}
\newcommand{\gsixMMissAllTwo}{13.42}
\newcommand{\gsixMMissLow}{7.0}
\newcommand{\gsixMMissHigh}{10.0}
\newcommand{\gsixMDev}{10.3}
\newcommand{\gsixWinSevFull}{4.04}
\newcommand{\gsixWinSevTD}{0.21}
\newcommand{\gsixAbsSrcLo}{33.0}
\newcommand{\gsixAbsSrcHi}{40.8}
\newcommand{\gsixAbsFullLo}{35.6}
\newcommand{\gsixAbsFullHi}{42.4}
\newcommand{\gsixAbsMaxGainCost}{0.291}
\newcommand{\gsixAbsAheadNames}{PolyU and SIDD}
\newcommand{\gsixWwTwoFull}{$-$0.64}
\newcommand{\gsixWwTwoTD}{+0.28}
\newcommand{\gsixWwTwoCamera}{Apple iPhone 6s}
\newcommand{\gsixWwTwoNet}{NAFNet}
\newcommand{\gsixGhSkipShare}{6.8}
\newcommand{\gsixGhSkipSevere}{21.4}
\newcommand{\gsixGhSkipGainFull}{0.31}
\newcommand{\gsixGhBestName}{DNI U}
\newcommand{\gsixFigMQHat}{0.25}
\newcommand{\gsixFigMMuHat}{0.66}
\newcommand{\gsixFigMCAmp}{0.50}
\newcommand{\gsixFcLearnedOne}{24{,}400}
\newcommand{\gsixFcLoraAboveNames}{DRUNet, NAFNet-32 and Restormer}

\ifdefined\gsixTCells\else\newcommand{\gsixTCells}{12}
\newcommand{\gsixTRealUnits}{278}
\newcommand{\gsixTRealFull}{9.34}
\newcommand{\gsixTRealTwo}{0.57}
\newcommand{\gsixTGaussFull}{4.49}
\newcommand{\gsixTGaussTwo}{0.29}
\fi
\providecommand{\gsixTRealUnits}{\gsixWRealAllN}\providecommand{\gsixTRealFull}{\gsixWRealAllFullHOne}\providecommand{\gsixTRealTwo}{\gsixWRealAllTwoHOne}
\providecommand{\gsixTGaussFull}{\gsixWGaussAllFullHOne}\providecommand{\gsixTGaussTwo}{\gsixWGaussAllTwoHOne}

\title{Shrink by the Tail, Cap by the Mean:\\Per-Crop Dosing for Few-Shot Denoiser Adaptation}
\author{Mohammed Wattad \qquad Tamir Shor \qquad Alexander M. Bronstein\\
Faculty of Computer Science, Technion -- Israel Institute of Technology, Haifa, Israel
}

\begin{document}
\raggedbottom
\twocolumn[{%
\renewcommand\twocolumn[1][]{#1}%
\maketitle
\begin{center}
\centering
\captionsetup{type=figure}
\includegraphics[width=\textwidth]{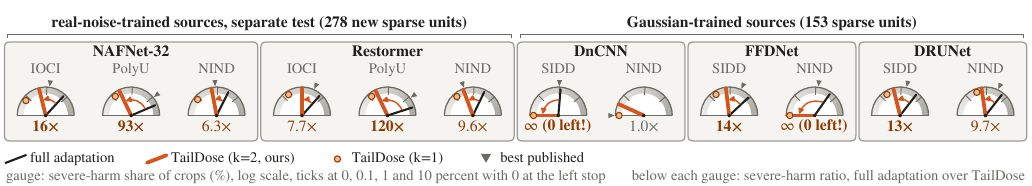}\makebox[0pt][r]{\makebox[\textwidth][l]{\raisebox{1.12in}{\small\textbf{(a)}}}}\\[1pt]
\includegraphics[width=\textwidth]{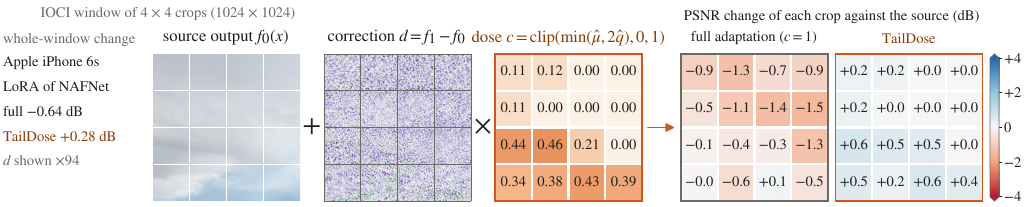}\makebox[0pt][r]{\makebox[\textwidth][l]{\raisebox{1.25in}{\small\textbf{(b)}}}}
\captionof{figure}{\textbf{\ours{} removes most severe harm of full adaptation (a) and can lift a harmed window back above the source (b).} (a) Per source network and dataset of the sparse adapters, the share of crops more than 1~dB worse than the source (log scale, left stop at zero) under full adaptation, \ours{} and \ours{} ($k=1$), their ratio below, and the best published controller (grey triangle, the least severe harm among those gaining at least as much as \ours{}, else the most gain). The real-noise group holds the \gsixTRealUnits{} new sparse adapters of a separate test of the frozen quantile model, on which \ours{} meets both criteria (\gsixTRealFull{} to \gsixTRealTwo{} percent), and the Gaussian group is descriptive (\gsixTGaussFull{} to \gsixTGaussTwo{} percent). (b) One IOCI window (\gsixWwTwoCamera{}, LoRA of \gsixWwTwoNet{}), chosen by a fixed rule to illustrate the effect (not representative), changes by \gsixWwTwoFull{}~dB under full adaptation and \gsixWwTwoTD{}~dB under \ours{}.}
\label{fig:teaser}
\end{center}%
\vspace{0.5em}
}]

\begin{abstract}
Adapting a pretrained denoiser to a new camera from four labeled pairs raises its mean PSNR but leaves many crops worse, and no label at deployment reveals which. Common controllers scale every correction by one shared amount, leaving harmed crops unprotected. We give each crop its own dose $c$, the fraction of the correction it receives. A dose harms a crop exactly when a hidden quantity, the alignment of the correction with the remaining error of the pretrained output, falls below $c/2$. \ours{} shrinks by the tail and caps by the mean: it sets $c$ to twice a cautious low forecast of the alignment, capped at a forecast of its mean, a plug-in of the best dose under a bound on the chance of harm. Harm then needs a forecast that is too high, so for any forecast the harmed share is at most its miss rate. Two small models, trained once on development cameras, serve all cameras and adapters. Across \gsixParNCameras{} cameras, against applying the correction everywhere, \ours{} makes \textbf{\gsixParLowRatio{}$\times$} fewer crops more than 1~dB worse on low-rank adapters and \textbf{\gsixParSpRatio{}$\times$} fewer on sparse adapters, with mean-gain changes of \gsixParLowdG{} and \gsixParSpdG{}~dB, neither reliably different from zero.
\end{abstract}

\section{Introduction}
\label{sec:intro}
\begin{figure*}[t]
\centering
\includegraphics[width=\textwidth]{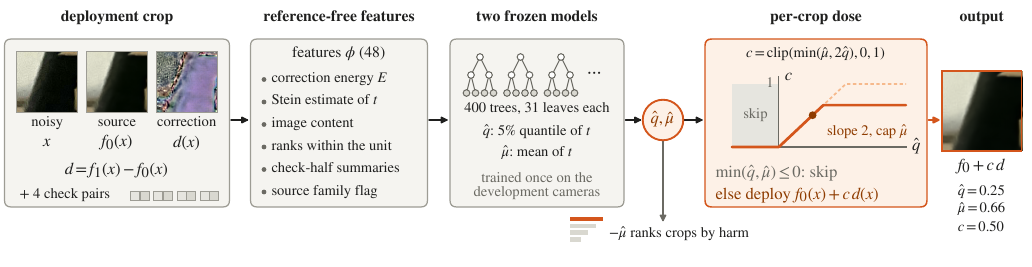}
\caption{\textbf{Two small frozen models set the dose of each crop.} Reference-free features, including summaries of the check halves, feed a quantile model and a mean model. Their forecasts set the dose $c=\clip(\min(\hat\mu,\gsixVThK{}\,\hat q),0,1)$, keep the source where either is at most zero, and $-\hat\mu$ ranks crops by harm. The example crop has $\hat q=\gsixFigMQHat{}$ and $\hat\mu=\gsixFigMMuHat{}$ and receives $c=\gsixFigMCAmp{}$.}
\label{fig:overview}
\end{figure*}
A pretrained denoiser shipped to a new camera can be adapted from a few labeled pairs, each a noisy shot and its clean reference, with a low-rank adapter (LoRA)~\cite{lora} or a sparse adapter that retrains a fixed subset of the existing weights~\cite{diffpruning,fishmask}. We call the pretrained network the \emph{source}, and applying the adapted network to every $\ciCrop{}\times\ciCrop{}$ image crop \emph{full adaptation}. Full adaptation raises the mean PSNR, but the mean nets the losses of some crops against the gains of others. We therefore evaluate adaptation per crop, next to the mean gain, with three harm measures: the \emph{harmed share} of crops worse than the source, the \emph{severe-harm share} of crops more than 1~dB worse (\emph{severe harm}), and the \emph{downside}, the mean PSNR shortfall below the source. Over our settings, full adaptation gained \gsixVFullAllG{}~dB yet harmed \gsixVFullAllH{} percent of crops. On the PolyU DSLRs~\cite{polyu}, it raised the mean PSNR by \gsixVFullPuG{}~dB, yet it left \gsixVFullPuH{} percent of crops worse than the source and \gsixVFullPuHOne{} percent severely harmed. Nothing available at deployment says which crops got worse, and the labeled pairs do not reveal it either. On their held-out halves, the only labeled data on a new camera, full adaptation harmed only \gsixrthChPuHoldHarmU{} percent of PolyU crops, less than half the share on deployed crops.

Regression-free updates are an established goal in vision~\cite{pct}, and an adapted restorer needs the same control for each crop. We call the difference between the adapted and the source outputs the \emph{correction}, and any rule that sets how much of it each crop receives a \emph{controller}. Most published controllers apply one shared coefficient: network interpolation~\cite{dni,wiseft} and Fisher merging~\cite{matena2022fisher} blend the two networks, and a fixed cap or halving~\cite{wattad2026tiny} shrinks every correction. SURE-LET-style mixing~\cite{surelet,mcsure} acts per crop under a Gaussian noise approximation. None of them bounds the share of crops that it harms.

Whether a scaled correction helps or hurts a crop is decided by a single scalar. With source output $f_0(x)$, correction $d$, unseen source error $r=y-f_0(x)$ for the clean crop $y$, and dose $c\in[0,1]$, a controller outputs $f_0(x)+c\,d$, and the \emph{alignment} $t=\ip rd/\norm d^2$ decides harm~\cite{wattad2026tiny}. The crop is harmed, $\norm{r-c\,d}>\norm r$, exactly when $c>0$ and $t<c/2$, so harm is a tail event of $t$ (it happens only when $t$ falls in its lower tail, below $c/2$). A dose of twice a forecast lower quantile $\hat q$ of $t$ harms a crop only where $t<\hat q$, that is, where the forecast misses. Harm control thus becomes a forecasting problem whose miss rate labeled development cameras can measure. Under a bound on the harm probability, the dose that maximises the expected reduction in squared error is twice the quantile capped at the mean alignment. We derive the factor two as the largest slope at which harm still requires a forecast miss, and it agrees with the classical safe range of linear shrinkage~\cite{jamesstein1961,hoerl1970}. Our step is to apply it to a per-crop forecast of the lower tail of $t$, made without labels of the test camera, which turns the safe range into a bound on harm.

Our controller \ours{} shrinks by the tail and caps by the mean. Two small companion models forecast the \hpAlphaPct\% quantile $\hat q$ and the mean $\hat\mu$ of $t$, and \ours{} sets the dose of each crop to the smaller of $\hat\mu$ and $2\hat q$, clipped to $[0,1]$. The quantile term confines harm to forecast misses, and the mean cap removes further severe harm. Their key input is a Stein estimate of $t$, which reads the noisy crop and the correction but never the clean image. We fit them once on development cameras, freeze them, and share them across every test camera, source network and adapter type (\cref{fig:overview}). \ours{} keeps the source where $\hat q\le0$ or $\hat\mu\le0$, and $-\hat\mu$ ranks crops by harm. \ours{} denotes this default, \ours{} ($k=2$), and the cap at $\hat q$ instead of $2\hat q$ gives the harm-averse setting \ours{} ($k=1$). Both leave the adapter unchanged and use no labels of the test camera beyond its four calibration pairs.

We evaluate \ours{} on two adapter types, \gsixParLowN{} low-rank and \gsixParSpN{} sparse \emph{units} over the same \gsixParNCameras{} cameras, \gsixParNSources{} source networks and \gsixNSettings{} \emph{settings}, where a unit is one adapter fitted on one draw of four labeled pairs and a setting is a camera dataset with the real-noise-trained or the Gaussian-trained sources. On low-rank adapters, \ours{} had a mean gain of \gsixVKTwoAllG{} against \gsixVFullAllG{}~dB for full adaptation, with \gsixVURatioFullAll{} times less downside and \gsixVHOneRatioFullAll{} times fewer severely harmed crops, from \gsixParLowFullHOne{} to \gsixParLowTDHOne{} percent (\cref{tab:persetting,tab:shrink} and the supplementary material). On sparse adapters, it gained \gsixParSpTDG{} against \gsixParSpFullG{}~dB, with \gsixParSpURatio{} times less downside and \gsixParSpRatio{} times fewer severely harmed crops, from \gsixParSpFullHOne{} to \gsixParSpTDHOne{} percent (\cref{tab:sparse,tab:shrink}). Neither gain difference is reliably different from zero, and on the sparse PolyU adapters, where full adaptation changed the mean PSNR by $\gsixSkSpFullPuG{}$~dB, \ours{} gained \gsixSkSpTDPuG{}~dB. In point estimate and per severely harmed percentage point, \ours{} gains more than every published controller and gate row in all four blocks of \cref{tab:main,tab:sparse}. Two separate tests, each with its method, units, endpoints and pass criteria fixed in a written plan before any test label was read, applied the frozen quantile model to new adapters and met their criteria, and \ours{} met both criteria on the same adapters.

\begin{table*}[t]
\centering
\fontsize{6.4}{7.7}\selectfont
\setlength{\tabcolsep}{0.9pt}
\caption{\gsixSpCaption{}}
\label{tab:main}
\begin{tabular*}{\textwidth}{@{\extracolsep{\fill}}l rrrrccc rrrrccc@{}}
\toprule
 & \multicolumn{7}{c}{\gsixSpBlockRe} & \multicolumn{7}{c}{\gsixSpBlockGa} \\
\cmidrule(lr){2-8}\cmidrule(l){9-15}
Controller & $G\uparrow$ & $U\downarrow$ & $H\downarrow$ & $H_1\downarrow$ & $G/H_1\uparrow$ & $U$ ratio$\uparrow$ & $H_1$ ratio$\uparrow$ & $G\uparrow$ & $U\downarrow$ & $H\downarrow$ & $H_1\downarrow$ & $G/H_1\uparrow$ & $U$ ratio$\uparrow$ & $H_1$ ratio$\uparrow$ \\
\midrule
Full adaptation & \gsixSpReFullG{} & \gsixSpReFullU{} & \gsixSpReFullH{} & \gsixSpReFullHOne{} & \gsixSpReFullGHOne{} & \gsixSpReURatioFull{} & \gsixSpReSevRatioFull{} & \gsixSpGaFullG{} & \gsixSpGaFullU{} & \gsixSpGaFullH{} & \gsixSpGaFullHOne{} & \gsixSpGaFullGHOne{} & \gsixSpGaURatioFull{} & \gsixSpGaSevRatioFull{} \\
\midrule
DNI G~\cite{dni} & \gsixSpReDniGG{} & \gsixSpReDniGU{} & \gsixSpReDniGH{} & \gsixSpReDniGHOne{} & \gsixSpReDniGGHOne{} & \gsixSpReURatioDniG{} & \gsixSpReSevRatioDniG{} & \gsixSpGaDniGG{} & \gsixSpGaDniGU{} & \gsixSpGaDniGH{} & \gsixSpGaDniGHOne{} & \gsixSpGaDniGGHOne{} & \gsixSpGaURatioDniG{} & \gsixSpGaSevRatioDniG{} \\
DNI U & \gsixSpReDniUG{} & \gsixSpReDniUU{} & \gsixSpReDniUH{} & \gsixSpReDniUHOne{} & \gsixSpReDniUGHOne{} & \gsixSpReURatioDniU{} & \gsixSpReSevRatioDniU{} & \gsixSpGaDniUG{} & \gsixSpGaDniUU{} & \gsixSpGaDniUH{} & \gsixSpGaDniUHOne{} & \gsixSpGaDniUGHOne{} & \gsixSpGaURatioDniU{} & \gsixSpGaSevRatioDniU{} \\
DNI CH & \gsixSpReDniCHG{} & \gsixSpReDniCHU{} & \gsixSpReDniCHH{} & \gsixSpReDniCHHOne{} & \gsixSpReDniCHGHOne{} & \gsixSpReURatioDniCH{} & \gsixSpReSevRatioDniCH{} & \gsixSpGaDniCHG{} & \gsixSpGaDniCHU{} & \gsixSpGaDniCHH{} & \gsixSpGaDniCHHOne{} & \gsixSpGaDniCHGHOne{} & \gsixSpGaURatioDniCH{} & \gsixSpGaSevRatioDniCH{} \\
SURE-LET-style crop~\cite{surelet,mcsure} & \gsixSpReSureCropG{} & \gsixSpReSureCropU{} & \gsixSpReSureCropH{} & \gsixSpReSureCropHOne{} & \gsixSpReSureCropGHOne{} & \gsixSpReURatioSureCrop{} & \gsixSpReSevRatioSureCrop{} & \gsixSpGaSureCropG{} & \gsixSpGaSureCropU{} & \textbf{\gsixSpGaSureCropH{}} & \gsixSpGaSureCropHOne{} & \gsixSpGaSureCropGHOne{} & \gsixSpGaURatioSureCrop{} & \gsixSpGaSevRatioSureCrop{} \\
SURE-LET-style image & \gsixSpReSureImgG{} & \gsixSpReSureImgU{} & \gsixSpReSureImgH{} & \gsixSpReSureImgHOne{} & \gsixSpReSureImgGHOne{} & \gsixSpReURatioSureImg{} & \gsixSpReSevRatioSureImg{} & \gsixSpGaSureImgG{} & \gsixSpGaSureImgU{} & \gsixSpGaSureImgH{} & \gsixSpGaSureImgHOne{} & \gsixSpGaSureImgGHOne{} & \gsixSpGaURatioSureImg{} & \gsixSpGaSevRatioSureImg{} \\
SURE-LET-style unit & \textbf{\gsixSpReSureUnitG{}} & \gsixSpReSureUnitU{} & \gsixSpReSureUnitH{} & \gsixSpReSureUnitHOne{} & \gsixSpReSureUnitGHOne{} & \gsixSpReURatioSureUnit{} & \gsixSpReSevRatioSureUnit{} & \gsixSpGaSureUnitG{} & \gsixSpGaSureUnitU{} & \gsixSpGaSureUnitH{} & \gsixSpGaSureUnitHOne{} & \gsixSpGaSureUnitGHOne{} & \gsixSpGaURatioSureUnit{} & \gsixSpGaSevRatioSureUnit{} \\
Fisher CH~\cite{matena2022fisher} & \gsixSpReFisherCHG{} & \gsixSpReFisherCHU{} & \gsixSpReFisherCHH{} & \gsixSpReFisherCHHOne{} & \gsixSpReFisherCHGHOne{} & \gsixSpReURatioFisherCH{} & \gsixSpReSevRatioFisherCH{} & \gsixSpGaFisherCHG{} & \gsixSpGaFisherCHU{} & \gsixSpGaFisherCHH{} & \gsixSpGaFisherCHHOne{} & \gsixSpGaFisherCHGHOne{} & \gsixSpGaURatioFisherCH{} & \gsixSpGaSevRatioFisherCH{} \\
Fixed radius cap~\cite{wattad2026tiny} & \gsixSpRePriorCapG{} & \gsixSpRePriorCapU{} & \gsixSpRePriorCapH{} & \gsixSpRePriorCapHOne{} & \gsixSpRePriorCapGHOne{} & \gsixSpReURatioPriorCap{} & \gsixSpReSevRatioPriorCap{} & \gsixSpGaPriorCapG{} & \gsixSpGaPriorCapU{} & \gsixSpGaPriorCapH{} & \gsixSpGaPriorCapHOne{} & \gsixSpGaPriorCapGHOne{} & \gsixSpGaURatioPriorCap{} & \gsixSpGaSevRatioPriorCap{} \\
Constant half~\cite{wattad2026tiny} & \gsixSpRePriorHalfG{} & \gsixSpRePriorHalfU{} & \gsixSpRePriorHalfH{} & \gsixSpRePriorHalfHOne{} & \gsixSpRePriorHalfGHOne{} & \gsixSpReURatioPriorHalf{} & \gsixSpReSevRatioPriorHalf{} & \gsixSpGaPriorHalfG{} & \gsixSpGaPriorHalfU{} & \gsixSpGaPriorHalfH{} & \gsixSpGaPriorHalfHOne{} & \gsixSpGaPriorHalfGHOne{} & \gsixSpGaURatioPriorHalf{} & \gsixSpGaSevRatioPriorHalf{} \\
\midrule
\multicolumn{15}{@{}l}{\emph{Label-free gate}} \\
Ckpt.\ disagreement G~\cite{deepens} & \gsixSpReCkdGG{} & \gsixSpReCkdGU{} & \gsixSpReCkdGH{} & \gsixSpReCkdGHOne{} & \gsixSpReCkdGGHOne{} & \gsixSpReURatioCkdG{} & \gsixSpReSevRatioCkdG{} & \gsixSpGaCkdGG{} & \gsixSpGaCkdGU{} & \gsixSpGaCkdGH{} & \gsixSpGaCkdGHOne{} & \gsixSpGaCkdGGHOne{} & \gsixSpGaURatioCkdG{} & \gsixSpGaSevRatioCkdG{} \\
Ckpt.\ disagreement U & \gsixSpReCkdUG{} & \gsixSpReCkdUU{} & \gsixSpReCkdUH{} & \gsixSpReCkdUHOne{} & \gsixSpReCkdUGHOne{} & \gsixSpReURatioCkdU{} & \gsixSpReSevRatioCkdU{} & \gsixSpGaCkdUG{} & \gsixSpGaCkdUU{} & \gsixSpGaCkdUH{} & \gsixSpGaCkdUHOne{} & \gsixSpGaCkdUGHOne{} & \gsixSpGaURatioCkdU{} & \gsixSpGaSevRatioCkdU{} \\
\midrule
\multicolumn{15}{@{}l}{\emph{Companion models alone and together}} \\
Mean only, $\hat\mu$ & \gsixSpReMeanMG{} & \gsixSpReMeanMU{} & \gsixSpReMeanMH{} & \gsixSpReMeanMHOne{} & \gsixSpReMeanMGHOne{} & \gsixSpReURatioMeanM{} & \gsixSpReSevRatioMeanM{} & \gsixSpGaMeanMG{} & \gsixSpGaMeanMU{} & \gsixSpGaMeanMH{} & \gsixSpGaMeanMHOne{} & \gsixSpGaMeanMGHOne{} & \gsixSpGaURatioMeanM{} & \gsixSpGaSevRatioMeanM{} \\
Quantile only, $2\hat q$ & \gsixSpReQOnlyG{} & \gsixSpReQOnlyU{} & \gsixSpReQOnlyH{} & \gsixSpReQOnlyHOne{} & \gsixSpReQOnlyGHOne{} & \gsixSpReURatioQOnly{} & \gsixSpReSevRatioQOnly{} & \textbf{\gsixSpGaQOnlyG{}} & \gsixSpGaQOnlyU{} & \gsixSpGaQOnlyH{} & \gsixSpGaQOnlyHOne{} & \gsixSpGaQOnlyGHOne{} & \gsixSpGaURatioQOnly{} & \gsixSpGaSevRatioQOnly{} \\
\ours{} ($k=\gsixVThKOne{}$) & \gsixSpReKOneG{} & \gsixSpReKOneU{} & \gsixSpReKOneH{} & \gsixSpReKOneHOne{} & \gsixSpReKOneGHOne{} & \gsixSpReURatioKOne{} & \gsixSpReSevRatioKOne{} & \gsixSpGaKOneG{} & \gsixSpGaKOneU{} & \gsixSpGaKOneH{} & \gsixSpGaKOneHOne{} & \gsixSpGaKOneGHOne{} & \gsixSpGaURatioKOne{} & \gsixSpGaSevRatioKOne{} \\
\textbf{\ours{} ($k=\gsixVThK{}$, ours)} & \gsixSpReKTwoG{} & \textbf{\gsixSpReKTwoU{}} & \textbf{\gsixSpReKTwoH{}} & \textbf{\gsixSpReKTwoHOne{}} & \textbf{\gsixSpReKTwoGHOne{}} &  &  & \gsixSpGaKTwoG{} & \textbf{\gsixSpGaKTwoU{}} & \gsixSpGaKTwoH{} & \textbf{\gsixSpGaKTwoHOne{}} & \textbf{\gsixSpGaKTwoGHOne{}} &  &  \\
\bottomrule
\end{tabular*}
\end{table*}

\paragraph{Contributions.}
\begin{enumerate}
\item \textbf{Harm bounded by forecast misses.} For any forecast, a dose of at most twice a forecast lower quantile of the alignment bounds the harm rate by the miss rate of the forecast, no larger slope does, and capping it at the mean alignment gives the best dose under a bound on the chance of harm (\cref{prop:harm,prop:cc,prop:suff}).
\item \textbf{Two small frozen models that shrink by the tail and cap by the mean.} \ours{} uses one quantile model and one mean model, trained once, to set, skip and rank crops on every test camera, source network and adapter type without refitting (\cref{sec:method}), and the mean cap gives the least severe harm of six dose rules in every dataset, on both adapter types (\cref{tab:shrink}).
\item \textbf{Evidence on low-rank and sparse adapters.} Averaged over the settings of each adapter type, \ours{} has a lower downside than every published controller with no reliable change in overall gain, none dominates it in point estimate, and it gains the most per severely harmed crop (\cref{tab:main,tab:sparse}).
\end{enumerate}

\section{Related work}
\label{sec:related}
\paragraph{Few-pair adaptation of denoisers.} LED~\cite{led}, ADL~\cite{adl}, S2R~\cite{s2r} and AINDNet~\cite{aindnet} adapt a denoiser to a new sensor, GainTuning~\cite{gaintuning} and ZS-N2N~\cite{zsn2n} adapt to each test image, and low-rank~\cite{lora} and sparse~\cite{diffpruning,fishmask} adapters limit the weights that change. They report mean quality, while we decide how much of a given adapter to apply on each crop.

\paragraph{Combining a source and an adapted model.} DNI~\cite{dni}, WiSE-FT~\cite{wiseft} and model soups~\cite{soups} move along the weight path with one coefficient, SURE-LET~\cite{surelet} and Monte-Carlo SURE~\cite{mcsure} mix denoisers by estimated risk, and Fisher merging~\cite{matena2022fisher} weights parameters by estimated information. DaWin~\cite{dawin} is the closest in spirit, since it sets a weight-interpolation coefficient per sample from the prediction entropies of the two models, a signal with no counterpart for a regression output. We instead forecast the lower tail of the alignment, the quantity that decides harm.

\paragraph{Regressions, selection and risk control.} Positive-congruent training~\cite{pct} and backward-compatible updates~\cite{trauble2021backward,srivastava2020backward} penalise regressions during training, and selective prediction~\cite{chow1970,selective} abstains. Quantile regression~\cite{koenker1978} estimates conditional tails, and conformal quantiles~\cite{cqr} and risk control~\cite{crc} certify them under exchangeability. Our harm bound needs no labeled data exchangeable with the test camera, because it holds crop by crop as an inclusion of events, not as a coverage statement.

\section{Harm is a tail event of one scalar}
\label{sec:problem}\label{sec:analysis}
\begin {table*}[t] \centering \fontsize {7.4}{8.8}\selectfont \setlength {\tabcolsep }{3.6pt} \caption {\textbf {\ours {} lowers the downside and the severe-harm share of full adaptation in every setting, harms fewer crops than its forecast misses, and its forecast ranks crops by harm and skips \gsixVSkipCatchSevereAll {} percent of the crops that full adaptation harms severely.} Ratios: downside $U$ and severe-harm share of full adaptation over those of \ours {}, with point estimate and camera-clustered \gsixBootLevel {} percent percentile interval (\gsixBootN {} replicates), Overall with equal setting weights. PolyU has \gsixYCamPu {} cameras and SIDD and NIND \gsixYCamSd {} each, so their per-setting intervals are coarse, and the Overall row resamples all \gsixYCamAll {} cameras. Harmed: share of crops worse than the source. Miss: share of crops whose alignment fell below the forecast ($t<\hat q$), the realised level of \cref {prop:suff}. By \cref {prop:harm}(iii) the harmed share of \ours {} is at most the miss share, and it is below it in every setting. Skip: share of crops with $c=0$, and share of the crops that full adaptation harms severely which \ours {} skips, pooled over crops in the Overall row. AUC: ranking of crops by severe harm under full adaptation with $-\hat \mu $ of the mean model, and the mean over settings in the Overall row. AUC any: the same ranking of crops by any harm. The AUCs of $-\hat q$ and of correction energy are in the supplementary material.} \label {tab:persetting} \begin {tabular*}{\textwidth }{@{\extracolsep {\fill }}lrrrrrrrrr@{}} \toprule & \multicolumn {2}{c}{Full over \ours {}, with interval} & \multicolumn {2}{c}{Harmed (\%)} & Miss (\%) & \multicolumn {2}{c}{Skip (\%)} & \multicolumn {1}{c}{AUC severe} & \multicolumn {1}{c}{AUC any} \\ \cmidrule (lr){2-3}\cmidrule (lr){4-5}\cmidrule (lr){6-6}\cmidrule (lr){7-8}\cmidrule (lr){9-9}\cmidrule (l){10-10} Setting & $U$ ratio$\uparrow $ & severe-harm ratio$\uparrow $ & full & \ours {} & $t<\hat q$ & crops & severe & $-\hat \mu $ & $-\hat \mu $ \\ \midrule IOCI & \gsixVURatioFullIo {} [\gsixVURatioFullIoLo {}, \gsixVURatioFullIoHi {}] & \gsixVHOneRatioFullIo {} [\gsixVHOneRatioFullIoLo {}, \gsixVHOneRatioFullIoHi {}] & \gsixVFullIoH {} & \gsixVKTwoIoH {} & \gsixMMissIoTwo {} & \gsixVSkipShareIo {} & \gsixVSkipCatchSevereIo {} & \gsixVAucSevereMIo {} & \gsixVAucAnyMIo {} \\ PolyU & \gsixVURatioFullPu {} [\gsixVURatioFullPuLo {}, \gsixVURatioFullPuHi {}] & \gsixVHOneRatioFullPu {} [\gsixVHOneRatioFullPuLo {}, \gsixVHOneRatioFullPuHi {}] & \gsixVFullPuH {} & \gsixVKTwoPuH {} & \gsixMMissPuTwo {} & \gsixVSkipSharePu {} & \gsixVSkipCatchSeverePu {} & \gsixVAucSevereMPu {} & \gsixVAucAnyMPu {} \\ SIDD & \gsixVURatioFullSd {} [\gsixVURatioFullSdLo {}, \gsixVURatioFullSdHi {}] & \gsixVHOneRatioFullSd {} [\gsixVHOneRatioFullSdLo {}, \gsixVHOneRatioFullSdHi {}] & \gsixVFullSdH {} & \gsixVKTwoSdH {} & \gsixMMissSdTwo {} & \gsixVSkipShareSd {} & \gsixVSkipCatchSevereSd {} & \gsixVAucSevereMSd {} & \gsixVAucAnyMSd {} \\ NIND, real-noise & \gsixVURatioFullNr {} [\gsixVURatioFullNrLo {}, \gsixVURatioFullNrHi {}] & \gsixVHOneRatioFullNr {} [\gsixVHOneRatioFullNrLo {}, \gsixVHOneRatioFullNrHi {}] & \gsixVFullNrH {} & \gsixVKTwoNrH {} & \gsixMMissNrTwo {} & \gsixVSkipShareNr {} & \gsixVSkipCatchSevereNr {} & \gsixVAucSevereMNr {} & \gsixVAucAnyMNr {} \\ NIND, Gaussian & \gsixVURatioFullNg {} [\gsixVURatioFullNgLo {}, \gsixVURatioFullNgHi {}] & \gsixVHOneRatioFullNg {} [\gsixVHOneRatioFullNgLo {}, \gsixVHOneRatioFullNgHi {}] & \gsixVFullNgH {} & \gsixVKTwoNgH {} & \gsixMMissNgTwo {} & \gsixVSkipShareNg {} & \gsixVSkipCatchSevereNg {} & \gsixVAucSevereMNg {} & \gsixVAucAnyMNg {} \\ \midrule Overall & \gsixVURatioFullAll {} [\gsixVURatioFullAllLo {}, \gsixVURatioFullAllHi {}] & \gsixVHOneRatioFullAll {} [\gsixVHOneRatioFullAllLo {}, \gsixVHOneRatioFullAllHi {}] & \gsixVFullAllH {} & \gsixVKTwoAllH {} & \gsixMMissAllTwo {} & \gsixVSkipShareAll {} & \gsixVSkipCatchSevereAll {} & \gsixVAucSevereMAll {} & \gsixVAucAnyMAll {} \\ \bottomrule \end {tabular*} \end {table*}

\paragraph{Setting.} A source denoiser $f_0$ is adapted from four labeled \emph{calibration pairs}, giving $f_1$. A test crop $x$ has $m$ output values, a clean reference $y$ unavailable at deployment, correction $d=f_1(x)-f_0(x)$ and source residual $r=y-f_0(x)$. A controller reads reference-free features $\phi(x)$ of $x$, $f_0(x)$, $d$ and the calibration pairs, and deploys
\begin{equation}
 f_{\dep}(x)=f_0(x)+c(\phi(x))\,d(x),\qquad c\in[0,1].
 \label{eq:deploy}
\end{equation}
Let $E=\norm d^2/m$ be the \emph{correction energy}, $R_0=\norm r^2/m$ and $t=\ip rd/\norm d^2$ the \emph{alignment}. Assume $E,R_0>0$ and that $\phi$ contains $E$. The reduction in mean squared error is
\begin{equation}
 D_c=\tfrac1m\big(\norm r^2-\norm{r-c\,d}^2\big)=E\,c\,(2t-c),
 \label{eq:scaled}
\end{equation}
and the PSNR change $\Delta_c=-10\log_{10}(1-D_c/R_0)$ has the sign of $D_c$. Wattad et al.~\cite{wattad2026tiny} used this expansion to analyse fixed halving and capping of MRI corrections. We evaluate a controller over the crops of a camera by the mean gain $G=\E[\Delta_c]$ and three harm measures, the harmed share $H=\Pr(\Delta_c<0)$, the severe-harm share $H_1=\Pr(\Delta_c<-1\,\mathrm{dB})$ and the downside $U=\E[(-\Delta_c)_+]$. Since $G=\E[(\Delta_c)_+]-U$, the mean gain nets the downside against the gains of helped crops, and the harm measures show what it hides. We also report the gain per harmed and per severely harmed percentage point, $G/H$ and $G/H_1$, as a ratio of the aggregate values, not a mean of per-unit ratios. Cautious rules score high on both, so each is read next to $G$. We write $\clip(v,0,1)=\min(\max(v,0),1)$.

\begin{proposition}[\PropHarmTitle]\label{prop:harm}
\PropHarmBody
\end{proposition}
\noindent\textbf{For \ours{}.} By (i), \ours{} harms a crop only when its alignment falls below half its dose. By (ii), no output dose in $[0,1]$, that of \ours{} included, harms a crop more than full adaptation does, so \textbf{\ours{} can only remove harm relative to full adaptation}. By (iii), its dose $\clip(\min(\hat\mu,2\hat q),0,1)$ is at most $2\hat q$, so it harms a crop only where the forecast missed (\cref{sec:predict}). Proof in the supplementary material.

\noindent Severe harm, a subset of harm, obeys the same inclusions, so an output controller is judged by the harm it removes at the gain it keeps. At $c=2\hat q$ the reduction is $D_c=4E\hat q\,(t-\hat q)$, which for $\hat q>0$ is zero exactly at $t=\hat q$, the \emph{\mbox{break-even} point}. By (iii), for any forecast every slope $k\le2$ in $c=\clip(k\hat q,0,1)$ confines harm to forecast misses. The slope $k=2$ is the largest such slope, the one whose break-even point is the forecast itself, so that harm and a miss coincide on crops whose dose is $2\hat q$ strictly between zero and one (see the supplementary material). Where the mean cap binds, harm needs $t<\hat\mu/2<\hat q$, so the cap can only lower the harm probability. Forecasting matters because the \emph{check halves}, held-out column halves of the calibration pairs that also select the training step of the adapter, can understate harm severely, as on PolyU (\cref{sec:intro}).

\begin{proposition}[\PropCCTitle]\label{prop:cc}\label{prop:frontier}
\PropCCBody
\end{proposition}
\noindent\textbf{For \ours{}.} If the chance of harm may be at most $\alpha$, the best dose is the smaller of twice the $\alpha$-quantile of the alignment, which holds the chance of harm at $\alpha$, and the mean alignment, the best dose on average, and when the spread of $t$ is large relative to its mean the quantile term is the smaller one. \ours{} plugs in its two forecasts, $\hat q$ at $\alpha=\hpAlphaPct\%$ and $\hat\mu$ (\cref{tab:shrink}). Proof in the supplementary material.

\noindent Because harm is a threshold event on one scalar, the constraint of \cref{prop:cc} has the standard deterministic equivalent $c\le2q_\alpha$ of an individual chance constraint~\cite{charnes1959}. On confident crops twice the quantile exceeds the mean, and there the cap at $\tb$, the unconstrained squared-error optimum, raises the expected reduction without adding harm (see the supplementary material). \ours{} applies this cap with its mean forecast, which is not exact, so there the cap trades a little gain for less severe harm (\cref{sec:shrink}).

\begin{proposition}[\PropSuffTitle]\label{prop:suff}
\PropSuffBody
\end{proposition}
\noindent\textbf{For \ours{}.} The forecaster is trained for $\alpha=\hpAlphaPct\%$, but on a new camera it misses at its own rate. By (i) that rate is measurable on any labeled camera, and by (ii) it bounds the harm rate of \ours{} on crops with a dose strictly between zero and one, with equality where the mean cap does not bind. \Cref{tab:persetting} checks this per setting, where the harmed share stayed below the miss share everywhere. Proof in the supplementary material.

\noindent For $k<2$ the dose $\clip(\min(\hat\mu,k\hat q),0,1)$ is at most that of \ours{} on every crop, so by \cref{prop:harm}(i) \ours{} ($k=1$) harms only crops with $t<\hat q/2$, and by \cref{prop:harm}(ii) only crops that \ours{} also harms, and by no more. The supplementary material analyses \ours{} ($k=1$) under a location-scale model.

\section{\ours{}: shrink by the tail, cap by the mean}
\label{sec:method}
\subsection{The dose and the skip rule}
\ours{} sets $c$ in \cref{eq:deploy} to
\begin{equation}
 c_{\mathrm{\ours}}(x)=\clip\big(\min(\hat\mu(\phi(x)),\,k\,\hat q(\phi(x))),0,1\big),
 \label{eq:qsafe}
\end{equation}
with $k=\gsixVThK{}$, where $\hat q$ is a frozen forecast of the $\alpha$-quantile of $t$ at $\alpha=\hpAlphaPct\%$, fitted by minimising the pinball loss $\rho_\alpha(v)=v\,(\alpha-\mathbf 1\{v<0\})$ of $t-\hat q(\phi)$~\cite{koenker1978}, and $\hat\mu$ is a frozen forecast of the mean of $t$, fitted by squared error. A forecast of the mean alone would aim at the squared-error optimum, whose harm probability a mean does not control, so \ours{} caps it at twice the lower quantile, and the pair is the plug-in of \cref{prop:cc}. The minimiser is a quantile of the pooled development law, so on a test camera the forecast is a quantile at its realised level $a(\phi)$, and \cref{prop:suff} turns this standard calibration gap into the harm probability. The harm-averse setting \ours{} ($k=\gsixVThKOne{}$) uses the same forecasts and confines harm to crops with $t<\hat q/2$. \ours{} \emph{skips} a crop (keeps the source) exactly where $\hat q\le0$ or $\hat\mu\le0$. Because harm requires $t<c/2$, a low forecast marks a crop as exposed to harm, and $-\hat\mu$ serves as the harm score. With constant spread under a location-scale model, the mean orders crops exactly as the harm probability of full adaptation does (see the supplementary material).

\subsection{Two small models serve every camera}
\label{sec:candidates}\label{sec:training}
The \gsixVModelFeatures{} reference-free features describe the correction energy, a Stein alignment estimate~\cite{stein1981}, noisy-image content, ranks within the unit, summaries of the check halves and the type of source network (Gaussian-trained or real-noise-trained). The quantile model and the mean model are histogram gradient boosting models~\cite{lightgbm,sklearn} of \gsixVModelTrees{} trees with at most \gsixVModelLeafCap{} leaves, the first trained with the pinball loss and the second with squared error, each fitted once on the pooled development rows of \gsixVModelTrainCams{} development cameras. Each holds \gsixFcLearnedOne{} learned numbers, its split thresholds and leaf values, and together they are smaller than the LoRA of \gsixFcLoraAboveNames{} (see the supplementary material). The supplementary material lists every feature, defines the alignment estimate and reports ablations of each feature group.

\subsection{Published controllers}
\label{sec:baselines}
We compare \ours{} with \gsixNPublished{} published controllers applied to the same adapter. DNI~\cite{dni} interpolates source and adapted weights with one coefficient, unchanged from its paper. DNI G and DNI U are \emph{development-matched}: their coefficient is fixed on development cameras to a gain or a downside target. DNI CH selects its coefficient on the check halves, and Fisher CH merges the two weight sets with diagonal Fisher information~\cite{matena2022fisher} and a check-half coefficient. SURE-LET-style mixing~\cite{surelet,mcsure} applies the risk-estimate principle to the output mix, minimising a Monte-Carlo SURE estimate under a Gaussian noise approximation per crop, image or unit. The fixed radius cap limits the correction to a calibration-sized radius, and the constant half applies $c=1/2$~\cite{wattad2026tiny}. A label-free gate scales each crop by $1/(1+\kappa V)$, with $V$ the disagreement of the corrections across stored checkpoints of the adapter, a standard measure of predictive uncertainty~\cite{deepens}, and $\kappa$ development-matched as for DNI G and DNI U.

\section{Less harm at the same overall gain}
\label{sec:main}\label{sec:protocol}

\begin{figure*}[t]
\centering
\includegraphics[width=\textwidth]{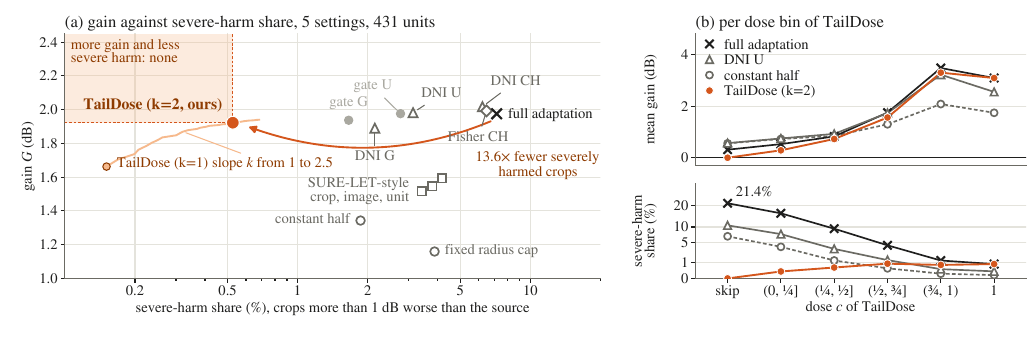}
\caption{\textbf{On low-rank adapters, no published controller has both more gain and less severe harm than \ours{}.} (a) Mean gain against severe-harm share over \gsixNUnits{} units: \ours{} has \gsixVHOneRatioFullAll{} times fewer severely harmed crops than full adaptation. (b) Mean gain and severe-harm share per dose bin of \ours{}, for full adaptation, \gsixGhBestName{} (best published by the rule of \cref{fig:teaser}), constant half (least severe harm among the published controllers) and \ours{}. On the \gsixGhSkipShare{} percent of crops that \ours{} skips, full adaptation gains \gsixGhSkipGainFull{}~dB and harms \gsixGhSkipSevere{} percent severely.}
\label{fig:gainharm}
\end{figure*}
\begin {table*}[t] \centering \fontsize {7}{8.4}\selectfont \setlength {\tabcolsep }{2.2pt} \caption {\textbf {On sparse adapters, \ours {} removes most severe harm of full adaptation, with no reliable change in gain.} Overall (\gsixBNSettings {} settings weighted equally): severe-harm ratio of full adaptation over \ours {} \gsixParSpRatio {} [\gsixParSpRatioLo {}, \gsixParSpRatioHi {}], downside ratio \gsixParSpURatio {}, gain change \gsixParSpdG {}~dB [\gsixParSpdGLo {}, \gsixParSpdGHi {}]. Real-noise block, the new adapters of a separate test of the frozen quantile model, on which \ours {} meets both criteria: severe-harm share of \ours {} minus full adaptation \gsixWRegCOne {} points [\gsixWRegCOneLo {}, \gsixWRegCOneHi {}], downside ratio of full adaptation over \ours {} \gsixWRegCTwo {} [\gsixWRegCTwoLo {}, \gsixWRegCTwoHi {}], gain difference (descriptive by design) \gsixWRegCThree {}~dB [\gsixWRegCThreeLo {}, \gsixWRegCThreeHi {}]. Gaussian-trained block: descriptive. Units weighted equally, $G$ and $U$ in dB, harmed ($H$) and severe-harm ($H_1$) shares in percent. $G/H_1$ is the gain per severely harmed percentage point of \cref {sec:problem}, in bold where best outside the $k=1$ row and highest for the harm-averse \ours {} ($k=1$). Published controllers and gate rows use the coefficients and $\kappa $ of \cref {tab:main}, fixed on development cameras.} \label {tab:sparse} \begin {tabular*}{\textwidth }{@{\extracolsep {\fill }}lrrrrrrrrrr@{}} \toprule & \multicolumn {5}{c}{Real-noise, new adapters of a separate test, \gsixWRealAllN {} units} & \multicolumn {5}{c}{Gaussian, descriptive, \gsixWGaussAllN {} units} \\ \cmidrule (lr){2-6}\cmidrule (l){7-11} Controller & $G\uparrow $ & $U\downarrow $ & $H\downarrow $ & $H_1\downarrow $ & $G/H_1\uparrow $ & $G\uparrow $ & $U\downarrow $ & $H\downarrow $ & $H_1\downarrow $ & $G/H_1\uparrow $ \\ \midrule Full adaptation & \gsixWRealAllFullG {} & \gsixWRealAllFullU {} & \gsixWRealAllFullH {} & \gsixWRealAllFullHOne {} & \gsixSrRealFullGHOne {} & \gsixWGaussAllFullG {} & \gsixWGaussAllFullU {} & \gsixWGaussAllFullH {} & \gsixWGaussAllFullHOne {} & \gsixSrGaussFullGHOne {} \\ \midrule DNI G & \gsixBRealDniGG {} & \gsixBRealDniGU {} & \gsixBRealDniGH {} & \gsixBRealDniGHOne {} & \gsixSrRealDniGGHOne {} & \gsixBGaussDniGG {} & \gsixBGaussDniGU {} & \gsixBGaussDniGH {} & \gsixBGaussDniGHOne {} & \gsixSrGaussDniGGHOne {} \\ DNI U & \gsixBRealDniUG {} & \gsixBRealDniUU {} & \gsixBRealDniUH {} & \gsixBRealDniUHOne {} & \gsixSrRealDniUGHOne {} & \gsixBGaussDniUG {} & \gsixBGaussDniUU {} & \gsixBGaussDniUH {} & \gsixBGaussDniUHOne {} & \gsixSrGaussDniUGHOne {} \\ DNI CH & \gsixBRealDniCHG {} & \gsixBRealDniCHU {} & \gsixBRealDniCHH {} & \gsixBRealDniCHHOne {} & \gsixSrRealDniCHGHOne {} & \gsixBGaussDniCHG {} & \gsixBGaussDniCHU {} & \gsixBGaussDniCHH {} & \gsixBGaussDniCHHOne {} & \gsixSrGaussDniCHGHOne {} \\ SURE-LET-style crop & \gsixBRealSureCropG {} & \gsixBRealSureCropU {} & \gsixBRealSureCropH {} & \gsixBRealSureCropHOne {} & \gsixSrRealSureCropGHOne {} & \gsixBGaussSureCropG {} & \gsixBGaussSureCropU {} & \gsixBGaussSureCropH {} & \gsixBGaussSureCropHOne {} & \gsixSrGaussSureCropGHOne {} \\ SURE-LET-style image & \gsixBRealSureImgG {} & \gsixBRealSureImgU {} & \gsixBRealSureImgH {} & \gsixBRealSureImgHOne {} & \gsixSrRealSureImgGHOne {} & \gsixBGaussSureImgG {} & \gsixBGaussSureImgU {} & \gsixBGaussSureImgH {} & \gsixBGaussSureImgHOne {} & \gsixSrGaussSureImgGHOne {} \\ SURE-LET-style unit & \gsixBRealSureUnitG {} & \gsixBRealSureUnitU {} & \gsixBRealSureUnitH {} & \gsixBRealSureUnitHOne {} & \gsixSrRealSureUnitGHOne {} & \gsixBGaussSureUnitG {} & \gsixBGaussSureUnitU {} & \gsixBGaussSureUnitH {} & \gsixBGaussSureUnitHOne {} & \gsixSrGaussSureUnitGHOne {} \\ Fisher CH & \gsixBRealFisherCHG {} & \gsixBRealFisherCHU {} & \gsixBRealFisherCHH {} & \gsixBRealFisherCHHOne {} & \gsixSrRealFisherCHGHOne {} & \gsixBGaussFisherCHG {} & \gsixBGaussFisherCHU {} & \gsixBGaussFisherCHH {} & \gsixBGaussFisherCHHOne {} & \gsixSrGaussFisherCHGHOne {} \\ Fixed radius cap & \gsixBRealPriorCapG {} & \gsixBRealPriorCapU {} & \gsixBRealPriorCapH {} & \gsixBRealPriorCapHOne {} & \gsixSrRealPriorCapGHOne {} & \gsixBGaussPriorCapG {} & \gsixBGaussPriorCapU {} & \gsixBGaussPriorCapH {} & \gsixBGaussPriorCapHOne {} & \gsixSrGaussPriorCapGHOne {} \\ Constant half & \gsixBRealPriorHalfG {} & \gsixBRealPriorHalfU {} & \gsixBRealPriorHalfH {} & \gsixBRealPriorHalfHOne {} & \gsixSrRealPriorHalfGHOne {} & \gsixBGaussPriorHalfG {} & \gsixBGaussPriorHalfU {} & \gsixBGaussPriorHalfH {} & \gsixBGaussPriorHalfHOne {} & \gsixSrGaussPriorHalfGHOne {} \\ Ckpt.\ disagreement G & \gsixBRealCkdGG {} & \gsixBRealCkdGU {} & \gsixBRealCkdGH {} & \gsixBRealCkdGHOne {} & \gsixSrRealCkdGGHOne {} & \gsixBGaussCkdGG {} & \gsixBGaussCkdGU {} & \gsixBGaussCkdGH {} & \gsixBGaussCkdGHOne {} & \gsixSrGaussCkdGGHOne {} \\ Ckpt.\ disagreement U & \gsixBRealCkdUG {} & \gsixBRealCkdUU {} & \gsixBRealCkdUH {} & \gsixBRealCkdUHOne {} & \gsixSrRealCkdUGHOne {} & \gsixBGaussCkdUG {} & \gsixBGaussCkdUU {} & \gsixBGaussCkdUH {} & \gsixBGaussCkdUHOne {} & \gsixSrGaussCkdUGHOne {} \\ \midrule \ours {} ($k{=}1$) & \gsixWRealAllOneG {} & \gsixWRealAllOneU {} & \gsixWRealAllOneH {} & \gsixWRealAllOneHOne {} & \gsixSrRealOneGHOne {} & \gsixWGaussAllOneG {} & \gsixWGaussAllOneU {} & \gsixWGaussAllOneH {} & \gsixWGaussAllOneHOne {} & \gsixSrGaussOneGHOne {} \\ \textbf {\ours {} ($k{=}2$)} & \gsixWRealAllTwoG {} & \gsixWRealAllTwoU {} & \gsixWRealAllTwoH {} & \gsixWRealAllTwoHOne {} & \textbf {\gsixSrRealTwoGHOne {}} & \gsixWGaussAllTwoG {} & \gsixWGaussAllTwoU {} & \gsixWGaussAllTwoH {} & \gsixWGaussAllTwoHOne {} & \textbf {\gsixSrGaussTwoGHOne {}} \\ \bottomrule \end {tabular*} \end {table*}

\begin{table*}[t]
\centering
\fontsize{7.4}{8.8}\selectfont
\setlength{\tabcolsep}{3.2pt}
\caption{\textbf{Capping twice the quantile at the mean gives the least severe harm of the six dose rules in every dataset and the most gain per harmed and per severely harmed crop overall, on both adapter types.} Each cell is the mean gain (dB) / severe-harm share (\%), on the same units for every rule, units weighted equally within a dataset and datasets equally overall, in bold where best in its column. $G/H$ and $G/H_1$ are the ratios of \cref{sec:problem} on the overall values, in bold where best. The quantile-only rule is \ours{} without the mean cap.}
\label{tab:shrink}
\begin{tabular*}{\textwidth}{@{\extracolsep{\fill}}lrrrrrrrr@{}}
\toprule
Dose rule & IOCI & PolyU & SIDD & NIND, real & NIND, Gauss. & Overall & $G/H\uparrow$ & $G/H_1\uparrow$ \\
\midrule
\multicolumn{9}{@{}l}{\emph{Low-rank adapters}} \\
Full adaptation, $c=1$ & \textbf{\gsixSkLoFullIoG{}} / \gsixSkLoFullIoHOne{} & \gsixSkLoFullPuG{} / \gsixSkLoFullPuHOne{} & \gsixSkLoFullSdG{} / \gsixSkLoFullSdHOne{} & \textbf{\gsixSkLoFullNrG{}} / \gsixSkLoFullNrHOne{} & \textbf{\gsixSkLoFullNgG{}} / \gsixSkLoFullNgHOne{} & \gsixSkLoFullAllG{} / \gsixSkLoFullAllHOne{} & \gsixSkLoFullAllGH{} & \gsixSkLoFullAllGHOne{} \\
Mean only, $\clip(\hat\mu,0,1)$ & \gsixSkLoMeanIoG{} / \gsixSkLoMeanIoHOne{} & \textbf{\gsixSkLoMeanPuG{}} / \gsixSkLoMeanPuHOne{} & \gsixSkLoMeanSdG{} / \gsixSkLoMeanSdHOne{} & \gsixSkLoMeanNrG{} / \gsixSkLoMeanNrHOne{} & \gsixSkLoMeanNgG{} / \gsixSkLoMeanNgHOne{} & \gsixSkLoMeanAllG{} / \gsixSkLoMeanAllHOne{} & \gsixSkLoMeanAllGH{} & \gsixSkLoMeanAllGHOne{} \\
Quantile only, $\clip(2\hat q,0,1)$ & \gsixSkLoQuantIoG{} / \gsixSkLoQuantIoHOne{} & \gsixSkLoQuantPuG{} / \gsixSkLoQuantPuHOne{} & \textbf{\gsixSkLoQuantSdG{}} / \gsixSkLoQuantSdHOne{} & \gsixSkLoQuantNrG{} / \gsixSkLoQuantNrHOne{} & \gsixSkLoQuantNgG{} / \gsixSkLoQuantNgHOne{} & \textbf{\gsixSkLoQuantAllG{}} / \gsixSkLoQuantAllHOne{} & \gsixSkLoQuantAllGH{} & \gsixSkLoQuantAllGHOne{} \\
\textbf{\ours{}}, $\clip(\min(\hat\mu,2\hat q),0,1)$ & \gsixSkLoTDIoG{} / \textbf{\gsixSkLoTDIoHOne{}} & \gsixSkLoTDPuG{} / \textbf{\gsixSkLoTDPuHOne{}} & \gsixSkLoTDSdG{} / \textbf{\gsixSkLoTDSdHOne{}} & \gsixSkLoTDNrG{} / \textbf{\gsixSkLoTDNrHOne{}} & \gsixSkLoTDNgG{} / \textbf{\gsixSkLoTDNgHOne{}} & \gsixSkLoTDAllG{} / \textbf{\gsixSkLoTDAllHOne{}} & \textbf{\gsixSkLoTDAllGH{}} & \textbf{\gsixSkLoTDAllGHOne{}} \\
Constant half, $c=1/2$~\cite{wattad2026tiny} & \gsixSkLoHalfIoG{} / \gsixSkLoHalfIoHOne{} & \gsixSkLoHalfPuG{} / \gsixSkLoHalfPuHOne{} & \gsixSkLoHalfSdG{} / \gsixSkLoHalfSdHOne{} & \gsixSkLoHalfNrG{} / \gsixSkLoHalfNrHOne{} & \gsixSkLoHalfNgG{} / \gsixSkLoHalfNgHOne{} & \gsixSkLoHalfAllG{} / \gsixSkLoHalfAllHOne{} & \gsixSkLoHalfAllGH{} & \gsixSkLoHalfAllGHOne{} \\
Fixed radius cap~\cite{wattad2026tiny} & \gsixSkLoCapIoG{} / \gsixSkLoCapIoHOne{} & \gsixSkLoCapPuG{} / \gsixSkLoCapPuHOne{} & \gsixSkLoCapSdG{} / \gsixSkLoCapSdHOne{} & \gsixSkLoCapNrG{} / \gsixSkLoCapNrHOne{} & \gsixSkLoCapNgG{} / \gsixSkLoCapNgHOne{} & \gsixSkLoCapAllG{} / \gsixSkLoCapAllHOne{} & \gsixSkLoCapAllGH{} & \gsixSkLoCapAllGHOne{} \\
\midrule
\multicolumn{9}{@{}l}{\emph{Sparse adapters}} \\
Full adaptation, $c=1$ & \textbf{\gsixSkSpFullIoG{}} / \gsixSkSpFullIoHOne{} & \gsixSkSpFullPuG{} / \gsixSkSpFullPuHOne{} & \gsixSkSpFullSdG{} / \gsixSkSpFullSdHOne{} & \textbf{\gsixSkSpFullNrG{}} / \gsixSkSpFullNrHOne{} & \gsixSkSpFullNgG{} / \gsixSkSpFullNgHOne{} & \gsixSkSpFullAllG{} / \gsixSkSpFullAllHOne{} & \gsixSkSpFullAllGH{} & \gsixSkSpFullAllGHOne{} \\
Mean only, $\clip(\hat\mu,0,1)$ & \gsixSkSpMeanIoG{} / \gsixSkSpMeanIoHOne{} & \textbf{\gsixSkSpMeanPuG{}} / \gsixSkSpMeanPuHOne{} & \gsixSkSpMeanSdG{} / \gsixSkSpMeanSdHOne{} & \gsixSkSpMeanNrG{} / \gsixSkSpMeanNrHOne{} & \gsixSkSpMeanNgG{} / \gsixSkSpMeanNgHOne{} & \gsixSkSpMeanAllG{} / \gsixSkSpMeanAllHOne{} & \gsixSkSpMeanAllGH{} & \gsixSkSpMeanAllGHOne{} \\
Quantile only, $\clip(2\hat q,0,1)$ & \gsixSkSpQuantIoG{} / \gsixSkSpQuantIoHOne{} & \gsixSkSpQuantPuG{} / \gsixSkSpQuantPuHOne{} & \textbf{\gsixSkSpQuantSdG{}} / \gsixSkSpQuantSdHOne{} & \gsixSkSpQuantNrG{} / \gsixSkSpQuantNrHOne{} & \textbf{\gsixSkSpQuantNgG{}} / \gsixSkSpQuantNgHOne{} & \textbf{\gsixSkSpQuantAllG{}} / \gsixSkSpQuantAllHOne{} & \gsixSkSpQuantAllGH{} & \gsixSkSpQuantAllGHOne{} \\
\textbf{\ours{}}, $\clip(\min(\hat\mu,2\hat q),0,1)$ & \gsixSkSpTDIoG{} / \textbf{\gsixSkSpTDIoHOne{}} & \gsixSkSpTDPuG{} / \textbf{\gsixSkSpTDPuHOne{}} & \gsixSkSpTDSdG{} / \textbf{\gsixSkSpTDSdHOne{}} & \gsixSkSpTDNrG{} / \textbf{\gsixSkSpTDNrHOne{}} & \gsixSkSpTDNgG{} / \textbf{\gsixSkSpTDNgHOne{}} & \gsixSkSpTDAllG{} / \textbf{\gsixSkSpTDAllHOne{}} & \textbf{\gsixSkSpTDAllGH{}} & \textbf{\gsixSkSpTDAllGHOne{}} \\
Constant half, $c=1/2$~\cite{wattad2026tiny} & \gsixSkSpHalfIoG{} / \gsixSkSpHalfIoHOne{} & \gsixSkSpHalfPuG{} / \gsixSkSpHalfPuHOne{} & \gsixSkSpHalfSdG{} / \gsixSkSpHalfSdHOne{} & \gsixSkSpHalfNrG{} / \gsixSkSpHalfNrHOne{} & \gsixSkSpHalfNgG{} / \gsixSkSpHalfNgHOne{} & \gsixSkSpHalfAllG{} / \gsixSkSpHalfAllHOne{} & \gsixSkSpHalfAllGH{} & \gsixSkSpHalfAllGHOne{} \\
Fixed radius cap~\cite{wattad2026tiny} & \gsixSkSpCapIoG{} / \gsixSkSpCapIoHOne{} & \gsixSkSpCapPuG{} / \gsixSkSpCapPuHOne{} & \gsixSkSpCapSdG{} / \gsixSkSpCapSdHOne{} & \gsixSkSpCapNrG{} / \gsixSkSpCapNrHOne{} & \gsixSkSpCapNgG{} / \gsixSkSpCapNgHOne{} & \gsixSkSpCapAllG{} / \gsixSkSpCapAllHOne{} & \gsixSkSpCapAllGH{} & \gsixSkSpCapAllGHOne{} \\
\bottomrule
\end{tabular*}
\end{table*}

\paragraph{Data.} The real-noise-trained sources NAFNet~\cite{nafnet} and Restormer~\cite{restormer} are adapted to the IOCI cameras~\cite{ioci}, the PolyU Canon 5D Mark II and Sony A7 II~\cite{polyu} and three NIND cameras~\cite{nind} (\emph{NIND, real-noise}). The Gaussian-trained sources DnCNN~\cite{dncnn}, FFDNet~\cite{ffdnet} and DRUNet~\cite{drunet} are adapted to the SIDD phones Pixel, Nexus 6 and LG-G4~\cite{sidd} and to the same three NIND cameras (\emph{NIND, Gaussian}). Each source is adapted with two adapter types that wrap the same layers with the same per-layer budget (see the supplementary material), a LoRA of rank \ciRank{} and a sparse adapter that retrains a fixed subset of the existing weights, both trained with reconstruction loss only. Each unit is scored on $\ciCrop{}\times\ciCrop{}$ crops of other scenes than its calibration pairs. For both adapter types, crops are averaged within a unit, units within a setting, and the \gsixNSettings{} settings with equal weight. On the low-rank grid, the source reached a mean crop PSNR of \gsixAbsSrcLo{} to \gsixAbsSrcHi{}~dB across the settings, and full adaptation \gsixAbsFullLo{} to \gsixAbsFullHi{}~dB.

\paragraph{Hyperparameters and the check halves.} Each calibration pair is split by columns. The left half trains the adapter for \ciSteps{} steps, and the right half, the check half, is never trained on. Of the source and the weights saved after \hpCkList{} steps, the unit deploys the one with the lowest mean squared error on its check halves.

\paragraph{Protocol.} Every model and coefficient was fitted on development cameras without test references, and every controller is applied to the same units and crops. Two separate tests applied the frozen quantile model to new adapters, \gsixTsOneN{} NIND low-rank adapters of the Gaussian-trained sources and \gsixTsTwoN{} sparse adapters of the real-noise-trained sources, and both met their criteria, a severe-harm share reliably below that of full adaptation and a downside ratio above two. On the same adapters, with the same endpoints and intervals, \ours{} met both criteria too, with downside ratios of \gsixTsOneMCTwo{} and \gsixTsTwoMCTwo{} (see the supplementary material).

\subsection{Low-rank adapters: less harm, same gain}
\label{sec:sets}\label{sec:grid}\label{sec:hinge}
Averaged over the settings, \ours{} gained \gsixVKTwoAllG{}~dB and full adaptation \gsixVFullAllG{}~dB (see the supplementary material), a difference not reliably different from zero (see the supplementary material). Split by source type and averaged over the settings of each block (\cref{tab:main}), \ours{} had a lower downside than every published controller, gate row and full adaptation with both kinds of source network, and also the lowest harmed and severe-harm shares with real-noise-trained sources. At this gain, \ours{} had \gsixVURatioFullAll{} times less downside and a \gsixVHOneRatioFullAll{} times lower severe-harm share than full adaptation, with intervals above one in every setting (\cref{tab:persetting}). \Cref{fig:gainharm} places every controller by gain and severe harm. In every setting \ours{} stayed within \gsixAbsMaxGainCost{}~dB of the gain of full adaptation, and in point estimate it gained more on \gsixAbsAheadNames{} (see the supplementary material).

On whole IOCI windows, full adaptation made \gsixWinSevFull{} percent more than 1~dB worse and \ours{} \gsixWinSevTD{} percent (see the supplementary material).

\paragraph{Against published controllers.} One controller \emph{dominates} another when it has both more gain and less downside. Averaged over the settings and in point estimate, \ours{} dominates \gsixVNDomPubAll{} of the \gsixrthNPub{} published controllers, and no published controller, gate row or full adaptation dominates \ours{}. The \gsixVNMoreGainMainAll{} rows with more gain, \gsixVExNames{}, both gate rows and full adaptation, gain at most \gsixVMoreGaindGMaxAll{}~dB more at no less than \gsixVMoreGainURatioMinAll{} times its downside and \gsixVMoreGainSevRatioMinAll{} times its severe-harm share. Every interval of the overall downside ratio and of the severe-harm ratio lies above one, for all \gsixYNCompRows{} rows (see the supplementary material), and \ours{} lowers the severe-harm share of full adaptation for every source network (see the supplementary material).

\subsection{Sparse adapters: less harm, same gain}
\label{sec:sparse}
Averaged over the \gsixBNSettings{} sparse settings, \ours{} gained \gsixParSpTDG{}~dB and full adaptation \gsixParSpFullG{}~dB, a difference not reliably different from zero, and \ours{} had \gsixParSpURatio{} times less downside and a \gsixParSpRatio{} times lower severe-harm share (\cref{tab:sparse}). On the real-noise-trained sources, it cut the severe-harm share of full adaptation from \gsixWRealAllFullHOne{} to \gsixWRealAllTwoHOne{} percent at a gain change of \gsixWRegCThree{}~dB [\gsixWRegCThreeLo{}, \gsixWRegCThreeHi{}] (\cref{fig:teaser}(a)).

\paragraph{Against published controllers.} Over the \gsixBNSettings{} sparse settings weighted equally, \ours{} dominated \gsixBPubNDom{} of the \gsixBNPub{} published controllers and \gsixBGateNDomWord{} of the two gate rows, every downside interval of these rows excluded zero, overall and in the real-noise block, and none of them had a lower downside in either block (\cref{tab:sparse} and the supplementary material).

\paragraph{Gain per harmed crop.} Per severely harmed percentage point, \ours{} gains more than every published controller, gate and full adaptation in all four blocks of \cref{tab:main,tab:sparse}, on low-rank adapters \gsixSpReKTwoGHOne{} and \gsixSpGaKTwoGHOne{}~dB against at most \gsixSpReGHOneBestComp{} and \gsixSpGaGHOneBestComp{}~dB. The advantage is reliably above one on sparse adapters with real-noise-trained sources, the separate test, with the best row chosen again per bootstrap replicate (see the supplementary material).

\subsection{Shrink by the tail, cap by the mean}
\label{sec:shrink}
\Cref{tab:shrink} is the ablation of \ours{}: six dose rules on the same units. A fixed amount gives up gain on well-aligned crops and still leaves harm: the constant half gained \gsixSkLoHalfAllG{} and \gsixSkSpHalfAllG{}~dB on low-rank and sparse adapters, with \gsixSkLoHalfAllHOne{} and \gsixSkSpHalfAllHOne{} percent of crops severely harmed. The mean forecast alone keeps the gain but leaves harm where it is optimistic, most of all on PolyU (\gsixSkLoMeanPuHOne{} and \gsixSkSpMeanPuHOne{} percent). The quantile alone confines harm to its misses, and capping it at the mean, the plug-in of \cref{prop:cc}, had the least severe harm of the six rules in every dataset on both adapter types, below the quantile alone by \gsixSkLodHOneAll{} [\gsixSkLodHOneAllLo{}, \gsixSkLodHOneAllHi{}] and \gsixSkSpdHOneAll{} [\gsixSkSpdHOneAllLo{}, \gsixSkSpdHOneAllHi{}] points overall, at a gain change of \gsixSkLodGAll{} [\gsixSkLodGAllLo{}, \gsixSkLodGAllHi{}] and \gsixSkSpdGAll{} [\gsixSkSpdGAllLo{}, \gsixSkSpdGAllHi{}]~dB. It also gained the most per harmed and per severely harmed point.

\subsection{One forecast skips and ranks}
\label{sec:predict}\label{sec:uses}
\paragraph{Ranking harm.} Read as a score, the mean forecast $-\hat\mu$ ranks crops by the harm that full adaptation would cause, with an AUC from \gsixVAucSevereMMin{} to \gsixVAucSevereMMax{} for severe harm and from \gsixVAucAnyMMin{} to \gsixVAucAnyMMax{} for any harm (\cref{tab:persetting}). In point estimate it ranks both better than correction energy, the training-free score of a magnitude rule, in \gsixVAucSevereMWins{} and \gsixVAucAnyMWins{} of the \gsixNSettings{} settings (see the supplementary material).

\paragraph{Harm follows the miss rate.} By \cref{prop:harm}(iii), every crop harmed by \ours{} has $t<\hat q$, and all \gsixVContGridNHarmed{} harmed crops of the low-rank grid satisfy it. Where the dose is $2\hat q$ strictly between zero and one, by \cref{prop:harm}(i) a crop was harmed exactly when the forecast missed, and where the mean cap binds, harm needed $t<\hat\mu/2$. The harmed share of \ours{} was therefore below the miss share of its forecast in every setting (\cref{tab:persetting}). On IOCI, SIDD and both NIND settings the forecast missed on \gsixMMissLow{} to \gsixMMissHigh{} percent of crops, at or below the \gsixMDev{} percent measured on development cameras held out one at a time, although IOCI and NIND contributed no development camera. On PolyU it missed on \gsixMMissPu{} percent, and \ours{} still removed the most downside there (\gsixVURatioFullPu{} times).

\section{Limitations and conclusion}
\label{sec:limits}\label{sec:conclusion}
\paragraph{Limitations.} \ours{} is evaluated on sRGB denoising with low-rank and sparse adapters on \gsixNCameras{} cameras, and the two NIND settings share images. Doses are set per $\ciCrop{}\times\ciCrop{}$ crop. Blending doses that differ between neighbouring crops is left open. \Cref{prop:cc} assumes the conditional law of the alignment, and $\alpha$ is a training target, so the deployed harm rate is bounded by the realised miss rate of the quantile forecast, which was highest on PolyU (\gsixMMissPu{} percent).

\paragraph{Conclusion.} For any forecast, a dose of at most twice a forecast lower quantile of the alignment bounds the harm rate by the miss rate of the forecast, and the cap at the mean alignment gives the best dose under a harm bound. \ours{} shrinks by the tail and caps by the mean, and had \gsixParLowRatio{} times fewer severely harmed crops than full adaptation on low-rank adapters and \gsixParSpRatio{} times fewer on sparse adapters, with no reliable change in overall gain. The propositions hold for any restorer with an additive correction under squared error, and we test them on denoising.

\paragraph{LLM Usage Statement}
Opus 5.5 supported literature review, coding verification, figure editing, word selection, and proofreading of this manuscript. The authors developed the research idea, planned and ran the experiments, analyzed the results, and take full responsibility for the content.

\FloatBarrier

\clearpage
\includepdf[pages=-]{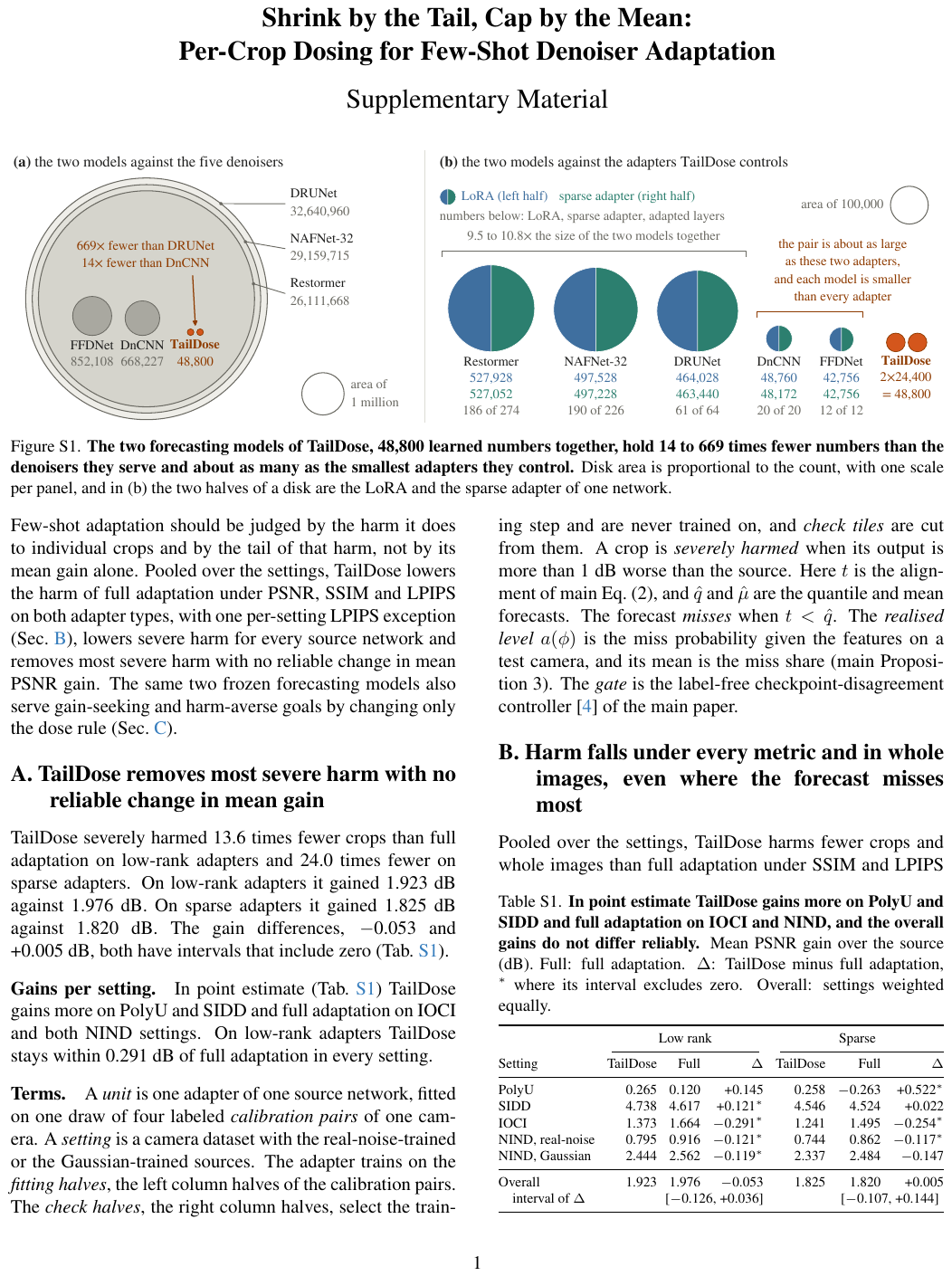}
\end{document}